\documentclass{article}

\usepackage[final]{corl_2022} %

\usepackage{graphicx}
\usepackage{natbib}
\usepackage{tikz}
\usepackage{comment}
\usepackage{amsmath,amssymb} %
\usepackage{color}
\usepackage{booktabs}
\usepackage{wrapfig}
\usepackage{booktabs}
\usepackage{multirow}
\usepackage{bm}
\usepackage{algorithm}
\usepackage{algorithmic}
\usepackage{microtype}
\usepackage{hyperref}

\hypersetup{
	colorlinks=true,
	urlcolor=magenta,
}
\DeclareMathOperator*{\argmin}{argmin} %

\usepackage[capitalize]{cleveref}
\crefname{section}{Sec.}{Secs.}
\Crefname{section}{Section}{Sections}
\Crefname{table}{Table}{Tables}
\crefname{table}{Tab.}{Tabs.}

\def\eg{\emph{e.g}.} 
\def\ie{\emph{i.e}.}

\def\model_name{CADSim}

\def \short {} %
\ifx \short \undefined

\newcommand{\cuthalfcaptionup}{}
\newcommand{\cuthalfcaptiondown}{}

\else

\newcommand{\cuthalfcaptionup}{\vspace*{-10pt}}
\newcommand{\cuthalfcaptiondown}{\vspace*{-10pt}}

\fi

\usepackage{colortbl}
\usepackage{color}
\newcommand{\mycite}[1]{[\citeauthor{#1},~\citeyear{#1}]}
\newlength\savewidth

\definecolor{grey}{rgb}{0.95,0.95,0.95}
\definecolor{ForestGreen}{RGB}{34,139,34}

\title{CADSim: Robust and Scalable in-the-wild 3D Reconstruction for Controllable Sensor Simulation}

\author{
	Jingkang Wang$^{1,2}$\quad Sivabalan Manivasagam$^{1,2}$\quad Yun Chen$^{1,2}$\quad Ze Yang$^{1,2}$ \\ \textbf{Ioan Andrei Bârsan}$^{1,2}$ \quad \textbf{Anqi Joyce Yang}$^{1,2}$\quad \textbf{Wei-Chiu Ma}$^{1,3}$\quad \textbf{Raquel Urtasun}$^{1,2}$ \vspace{0.05in}\\
	Waabi$^{1}$ \quad {University of Toronto$^{2}$ \quad Massachusetts Institute of Technology$^{3}$}  \\
	{\tt\footnotesize{\{jwang,smanivasagam,zyang,ychen,abarsan,jyang,wma,urtasun\}@waabi.ai}} \\  %
	\url{https://waabi.ai/cadsim/} \vspace{-0.22in}
}

\begin{document}
\maketitle

\begin{abstract}
Realistic simulation is key to enabling safe and scalable development of %
self-driving vehicles.
A core component is simulating the sensors so that the entire
autonomy system
can be tested
in simulation.
Sensor simulation involves modeling traffic participants, such as vehicles, with high quality
appearance and articulated geometry,
and rendering them in real time.
The self-driving industry has typically employed artists to build these assets. However, this is expensive, slow, and may not reflect reality.
Instead, reconstructing
assets automatically from sensor data collected in the wild would provide a
better path to generating a diverse and large set
with good real-world coverage.
Nevertheless, current reconstruction approaches
struggle on in-the-wild sensor data, due to its  sparsity and noise.
To tackle these issues, we present \textit{\model_name},
which combines part-aware
object-class priors via a small set of CAD models with
differentiable rendering to
automatically reconstruct
vehicle geometry, including articulated wheels,
with high-quality appearance.
Our experiments show our method recovers more accurate shapes from sparse data
compared to existing approaches.
Importantly, it also trains and renders efficiently.
We demonstrate
our reconstructed vehicles
in several
applications, including  accurate testing %
of autonomy perception systems.
\end{abstract}

\keywords{3D Reconstruction, CAD models, Sensor Simulation, Self-Driving}

\begin{figure}[htbp!]
	\cuthalfcaptionup
	\begin{center}
		\includegraphics[width=0.9\textwidth]{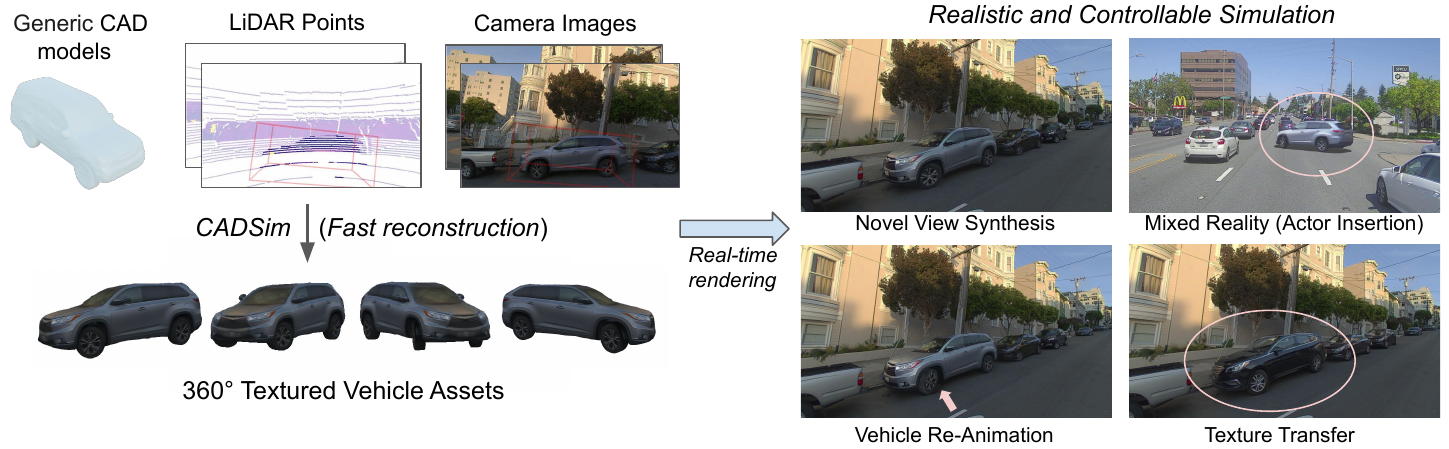}
	\end{center}
	\cuthalfcaptionup
	\vspace{-0.05in}
	\caption{\textbf{CADSim recovers shape, appearance and illumination in a robust and scalable way from sensor observations}. The reconstructed assets are of high fidelity, part-aware, geometry-aligned and compatible with graphics engine, enabling efficient, realistic and controllable simulation.
	}
	\cuthalfcaptiondown
	\label{fig:teaser}
\end{figure}

\section{Introduction}

Robots, such as self-driving vehicles (SDVs), learn to navigate environments and interact safely with other agents through experience.
It is critical for robots to %
handle rare or safety-critical situations which are challenging or dangerous to observe in the real world~\cite{tan2021scenegen}.
Simulation provides a solution for the robot to efficiently experience edge-case scenarios for
evaluation in a safe and cost-effective manner.
For proper testing that covers the full space of possible scenarios, the simulator should have a diverse and large set of traffic participants.
For example, vehicles have a wide variety of shapes, sizes, and appearances.
We want to ensure that the perception and autonomy systems can detect them and act appropriately  in all possible situations.
Existing self-driving simulators have a limited set of objects, as each 3D asset is typically designed manually, with fully specified shape, appearance, etc.
This is a time-consuming and expensive process that does not scale.
In this work, we propose to leverage real sensor data collected by the SDV in the wild to automatically build an asset library for realistic sensor simulation.
In-the-wild data is cost-efficient and scalable to obtain, enabling the collection of a large and diverse set of objects.

Specifically, given a sequence of
sensor data (camera images and LiDAR point clouds) collected
by an SDV, our goal is to automatically create digital replicas for all nearby vehicles in the scene and directly use them in a high-fidelity simulator to create many scenarios that never existed before.
To achieve high-quality and scalable sensor simulation, the reconstructed assets should (i) have precise shape and photorealistic appearance; (ii) be easily editable  %
to create new variations from the original assets, (iii) be controllable, such
as the ability to manipulate the vehicle to generate new behaviors (\eg, wheels
steering) and (iv) allow for real-time rendering, which is
important for closed-loop training and large-scale evaluation \cite{shacklett2021large}. Out of past
works using various 3D representations for reconstruction, such as point
clouds
\cite{yuan2018pcn,insafutdinov2018unsupervised,lin2018learning,gu2020weakly},
voxels %
\cite{wu2016learning,wu2017marrnet,wu2018learning},
or neural implicit representations \cite{mescheder2019occupancy,xu2019disn,lin2020sdf,duggal2022secrets}, the mesh representation best achieves the desired properties~\cite{sanders2016introduction,blender}.
In this paper we utilize mesh representations since they are widely used throughout
simulation~\cite{dosovitskiy2017carla,chen2021geosim,tu2020physically} and content creation
systems~\cite{liu2018paparazzi,Liu:CubicStyle:2019}, are easy to manipulate \cite{yifan2020neural,jakab2021keypointdeformer},
suitable for rigging and animation \cite{SMPL:2015,zuffi20173d,SMPL-X:2019,yang2020recovering},
support texturing and material properties \cite{kato2018neural,ravi2020accelerating},
and are fast to render \cite{foley1996computer}.

Generating meshes of nearby objects from
in-the-wild sensor data
is a challenging task.
The sensor data is sparse ($< 30$ views) and may be noisy. Furthermore, the objects of interest are observed from limited viewpoints and portions of the objects are unobserved, whereas we want to reconstruct the complete shape for simulation.
Previous mesh reconstruction methods tackle this task as an energy minimization problem, where given an initial sphere \cite{zhang2021ners} or ``mean-shape'' mesh template \cite{engelmann2017samp, kanazawa2018learning}, the goal is to optimize the mesh to be consistent with sensor data, subject to some regularization.
However, when operating on in-the-wild sensor data, these methods generate overly smoothed meshes that are inaccurate \cite{zhang2021ners, kanazawa2018learning}, or have severe artifacts and self-intersecting edges \cite{nicolet2021large}.
We demonstrate in our experiments that this is due to poor initialization.
Moreover, these approaches generate rigid meshes that cannot be controlled and lack articulation.
We therefore propose a simple and effective solution---to leverage class-specific CAD models to provide priors for reconstruction.
Not only do CAD models have high quality shape, they encode rich semantic information, such as where the wheels in a vehicle are and how they move. This enables better initialization and reconstruction of high-quality \emph{articulated} meshes from real-world data.

CAD models alone are not sufficient for realistic sensor simulation, as we we want to create digital twins of any possible vehicle.  Instead, we optimize our vehicle CAD model representation with a state-of-the-art differentiable renderer in an energy minimization framework.
This allows us to faithfully reconstruct
a diverse set of real-world vehicles, including their moving parts (\eg, wheels), from very sparse and noisy observations.
We compare against a wide range of reconstruction methods, including recent neural rendering-based approaches, and show improved reconstruction performance on both camera and LiDAR sensor data.
We show that our reconstructed vehicles can be used for a variety of applications such as novel view synthesis, texture transfer, mixed reality via injecting simulated actors into real data, and vehicle re-animation (Fig.\ \ref{fig:teaser}).
We also demonstrate that our approach creates a more realistic simulation environment for robots such as SDVs by evaluating perception models on simulated sensor data and showing small domain gap compared to real data.

\section{Related Work}

\paragraph{3D Representations for Sensor Simulation.} Triangular meshes are currently the de facto standard in real-time
graphics~\cite{sanders2016introduction,blender,dosovitskiy2017carla}.
While fast, flexible, compact, and widely adopted in game engines and %
simulators~\cite{dosovitskiy2017carla}, mesh-based assets are usually regarded as expensive to create due to the
need for human artists.
Surfels~\cite{pfister2000surfels} have also been leveraged in LiDAR simulation for robotic applications~\cite{yang2020surfelgan,manivasagam2020lidarsim}. However, their lack of geometric structure
makes them difficult to animate, edit, and compress.
Recently, volumetric representations have experienced a surge in popularity~\cite{sitzmann2021awesome,tewari2022advances,xie2022neural} thanks to their simplicity, differentiability, and flexibility in modeling non-Lambertian effects~\cite{mildenhall2020nerf,sitzmann2020implicit,boss2022samurai}.
These works can roughly be divided into neural volumetric rendering, such as NeRFs~\cite{mildenhall2020nerf,zhang2020nerf++,nerfactor,tancik2021learned,muller2022instant},
and implicit surfaces such as signed distance fields~\cite{park2019deepsdf,wang2021neus,zhang2021physg} or occupancy
fields~\cite{mescheder2019occupancy,yariv2020multiview,oechsle2021unisurf}.
Various methods have been proposed to speed up surface
and volumetric rendering~\cite{chabra2020deep,takikawa2021neural,hedman2021baking,yu2021plenoctrees,yu2021plenoxels}. However, these approaches still lag behind traditional rasterization pipeline in terms of rendering speed and throughput.
In our work, we use triangular meshes optimized from CAD models for efficient and accurate geometry modelling.

\paragraph{Multi-View 3D Reconstruction in the Wild.} Despite the incredible success achieved by recent neural volumetric representations, these works
are mainly demonstrated on synthetic datasets with hundreds of (noise-free) images and dense coverage on the camera viewpoints~\cite{mildenhall2020nerf,niemeyer2022regnerf,zhang2021ners}.
However, the real world input is usually much sparser (\eg, few views of a single object with similar view points)
and noisier (\eg, localization, calibration error and LiDAR noise) for real world applications such as robotics and self-driving. To address this issue,
recent works~\cite{chibane2021stereo,chen2021mvsnerf,trevithick2021grf} proposed conditional models that require expensive pre-training on other scenes with multi-view images and camera pose annotations.
Another recent line of work proposes learning mesh representations directly, either by deforming a generic mesh~\cite{loper2014opendr,kato2018neural,chen2019dibr,nicolet2021large,chen2021dibr++,monnier2022share}, or by leveraging a differentiable mesh extraction algorithm on top of an implicit representation~\cite{guo2020object,shen2021deep,munkberg2021extracting}.
Different from past works, we show that simple mesh optimization with differentiable rendering is sufficient for high-fidelity reconstruction.

\paragraph{CAD Priors.} Retrieving a 3D model conditioned on a single observation has been shown to be a strong baseline
which outperforms more sophisticated early approaches to single-view reconstruction~\cite{tatarchenko2019single}.
This motivates the use of imperfect but
readily available 3D CAD models as a powerful prior for
reconstruction\textcolor{blue}~\cite{engelmann2016joint,engelmann2017samp,chabot2017deep,wang2020directshape,koestler2020learning,jiang2020shapeflow,uy2020deformation,uy2021joint,lu2020permo}.
\cite{engelmann2016joint,engelmann2017samp,wang2020directshape} use the truncated SDF volumes to build the latent space for the CAD models for joint pose and shape estimation. %
\citet{uy2020deformation,uy2021joint} encode sparse observations and use them to retrieve the
most similar candidate from a CAD asset bank, which is then deformed to match the input.
While promising, none of the approaches in this area have been demonstrated
to work outside of synthetic~\cite{chang2015shapenet} datasets.
Most do not model realistic texture, nor are they part-aware.
\citet{lu2020permo} conduct a two-stage CAD-based deformation and handle the wheels separately. However, they recover the appearance only from single images without physics based materials and lighting models, and do not optimize the articulation from data.

\section{Method}

\begin{figure}[t]
    \begin{center}
        \includegraphics[width=0.95\textwidth]{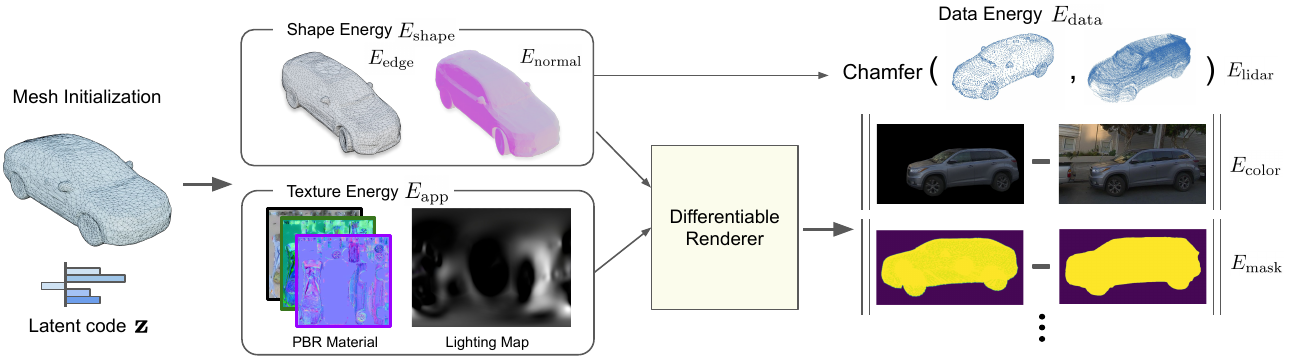}
    \end{center}
    \cuthalfcaptionup
	\vspace{-0.05in}
    \caption{Overview of \model_name{} for 3D Reconstruction on in-the-wild data. %
    }
    \cuthalfcaptiondown
    \vspace{-0.05in}
    \label{fig:overview}
\end{figure}

Given camera images and/or LiDAR point clouds
collected by a self-driving car, our goal is to automatically create digital replicas for all nearby vehicles in the scene and use them in a high-fidelity simulator.
We build our model based on the observation that existing CAD models are equipped with detailed geometry and animatable parts, which can be used as a prior during the reconstruction process.
Towards this goal, we propose an energy-based formulation that exploits CAD models, as well as visual and geometric cues from images and/or LiDAR %
for 3D reconstruction (Fig.~\ref{fig:overview}).
{We first describe our model representation and how it leverages CAD priors (Sec.~\ref{sec:cad_priors}).
Then we present our energy-based model (Sec.~\ref{sec:energy_formulation}), and how we conduct inference to generate the
vehicle mesh (Sec.~\ref{sec:inference}).}

\subsection{Model Representation}
\label{sec:cad_priors}

To faithfully simulate an object, %
we need to recover not only its geometry, but also its material.
Towards this goal, we exploit \emph{textured meshes} as underlying representation since they best meet the desired properties of modern high-fidelity simulators, by being efficient to animate, edit, and render.

\paragraph{Mesh representation.} A mesh $\mathcal{M} = (\mathbf{V}, \mathbf{F})$ is composed of a set of vertices $\mathbf V \in \mathbb{R}^{\rvert V \lvert\times3}$ and faces $\mathbf F \in \mathbb{N}^{\rvert F \lvert \times 3}$, where the faces define the connectivity of the vertices.
Our goal is to deform a mesh to match the observations from the sensor data. %
Typically, during deformation, the topology (\ie,  connectivity) of the mesh is fixed and only the vertices are ``moving'' \cite{SMPL:2015,kanazawa2018learning,goel2020shape,yifan2020neural}.
This strategy greatly simplifies the deformation process,
yet at the same time constrains its representation power.
For instance, if the original mesh topology is relatively simple (\eg, a sphere \cite{goel2020shape} or a ``mean-shape'' mesh template \cite{kanazawa2018learning}) and non-homeomorphic to the object of interest, the mesh may struggle to capture the fine-grained geometry (\eg, side mirrors of a car).

\paragraph{CAD models as priors.} To circumvent such limitations, we propose to incorporate shape priors from CAD models into the mesh reconstruction.
One straightforward approach is to directly exploit the CAD model to initialize the mesh $\mathcal{M}_{\text{CAD}} = (\mathbf{V}_{\text{CAD}}, \mathbf{F}_{\text{CAD}})$.
Since CAD models, by design, respect the topology of real-world objects, the mesh will be able to model finer details.
One obvious caveat, however, is that there is no structure among the vertices. Each vertex may move freely in the 3D space during the optimization;
 \eg{}, there is no guarantee that vertices belonging to a wheel will continue to form a cylinder.
We thus further incorporate the \emph{part information} from CAD models into the parameterization. This allows us to \emph{inherently} preserve part geometry, as well as
enable part manipulation (\eg, changing the steering of the wheels).

\begin{wrapfigure}{R}{0.6\textwidth}
	\vspace{-0.23in}
	\begin{center}
		\includegraphics[width=0.6\textwidth]{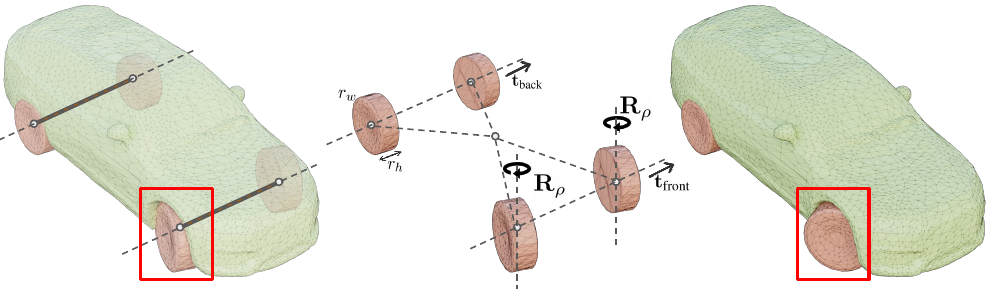}
	\end{center}
	\cuthalfcaptionup
	\vspace{-0.02in}
	\caption{\textbf{Vehicle model used in \model_name}. For simplicity, we only show the offsets along the axle axis.
	}
	\vspace{-0.05in}
	\cuthalfcaptiondown
	\label{fig:vehicle_model}
\end{wrapfigure}

We leverage the semantic part information from CAD models to partition the vehicle mesh $\mathcal{M}_{\text{CAD}}$ into a vehicle body $\{\mathbf V_{\mathrm{body}}, \mathbf{F}_{\mathrm{body}}\}$ and  wheels $\{\mathbf V_{\mathrm{wheel}}^{(k)}, \mathbf{F}_{\mathrm{wheel}}^{(k)}\}_{i=1}^{K}$. $K$ indicates the number of wheels a vehicle has. $K = 4$ for most vehicles but may be $6$ or more for large trucks.
We employ the fact that  the wheels  are usually the same within the same vehicle, and model all  wheels with the same underlying mesh $\{\mathbf V_{\mathrm{wheel}}, \mathbf{F}_{\mathrm{wheel}}\}$ and their individual relative pose $\mathbf T^{k}$ to the vehicle origin.
As there are a wide variety of vehicles with different wheel sizes and different relative positions to the vehicle body, we further add a scale factor
$\mathbf r = [r_w, r_h, r_w ]$
(wheel radius and thickness),
and per-axle translation offsets $\mathbf t_{\mathrm{front}}, \mathbf t_{\mathrm{back}} \in \mathbb R^{3}$ with respect to the wheel origin.
We can thus drastically reduce the degrees of freedom, enforce the shape of the wheel, and guarantee the relative positions of the wheels to be symmetric.
Importantly, the front-axle wheels can be steered and do not necessarily align with the body. We hence parameterize the front wheels to have a yaw-relative orientation $\rho$. The full vehicle mesh model (see Figure~\ref{fig:vehicle_model}) can thus be written as:
\begin{align}
\begin{split}
    \mathbf{V}_\mathrm{wheel}^{k}(\mathbf r, \mathbf{t}_{\text{front}}, \rho; \mathbf{V}_\mathrm{wheel}) = \mathbf T^\text{k} (\mathbf{R}_\rho\mathbf r\mathbf{V}_\mathrm{wheel}+\mathbf{t_\text{front}}) \quad k \in \{1 ,2\},&\quad\text{(front wheels)}\\
    \mathbf{V}_\mathrm{wheel}^{k}(\mathbf r, \mathbf{t}_{\text{back}}; \mathbf{V}_\mathrm{wheel}) = \mathbf T^\text{k} (\mathbf r\mathbf{V}_\mathrm{wheel}+\mathbf{t_\text{back}}) \quad k \in \{3 ,4, ..., K\},&\quad\text{(remaining wheels)}\\
    \mathbf{V} = \{\mathbf V_{\text{body}}, \mathbf{V}_\mathrm{wheel}^{(k)}\}  \quad k \in \{1, \dots, K\},&
\label{eq:cad}
\end{split}
\end{align}
with $\mathbf V_{\text{body}},\mathbf r, \mathbf{t}_{\text{back}}, \mathbf{t}_{\text{front}}, \rho$ being the free variables.

\paragraph{Extending to a CAD library.}
While a CAD model provides a useful prior for mesh initialization, it is likely that a single template mesh may not sufficiently cover a wide range of objects.
We thus resort to a CAD model library that consists of various vehicle types.
Inspired by \cite{engelmann2017samp,lu2020permo}, we represent a large variety of vehicles with a compact, low dimensional code, which allows us to quickly
determine the best CAD model to initialize from.
We first parameterize a collection of the CAD models with Eq.~\eqref{eq:cad}, simplify them so that they have
the same number of vertices and topology, and align them in a shared coordinate space with dense correspondences.
Finally, we apply principal component analysis on vertex coordinates {to obtain a shared low-dimensional code $\mathbf z$}.
Note that since all CAD models are parameterized with Eq.~\eqref{eq:cad}, the deformed mesh will reflect this
part structure as well.
{Please refer to the supp.\ material for more details.}

\paragraph{Appearance representation.}
In addition to geometry, our mesh representation must accurately capture how light interacts with its surface so that we can realistically reproduce sensor observations. Towards this goal, we parameterize the mesh appearance using a physically-based appearance representation.
Specifically, we represent appearance using a micro-facet BRDF model with the differentiable split sum environment lighting~\cite{karis2013real} proposed by~\citet{munkberg2021extracting}.
Since the topology of the mesh is fixed, we can build a one-to-one mapping relationship between each point on the mesh surface and each point in the 2D $(u, v)$ space.
We use a physically-based (PBR) material model from Disney~\cite{mcauley2012practical}, which contains a diffuse lobe $\mathbf k_d$ with an isotropic, specular GGX lobe~\cite{walter2007microfacet}.

\subsection{Energy Formulation}

\label{sec:energy_formulation}
The next step after the CAD model initialization is to optimize both the geometry and the appearance of the mesh such that it best matches the captured sensor data.
Our approach builds upon recent success on differentiable rendering \cite{kato2018renderer,liu2019softras,chen2019dibrender,ravi2020pytorch3d},
leveraging a differentiable renderer that takes as input variables such as the sensor pose and our textured mesh representation, and outputs a realistic simulation of the object.
Differentiable rendering allows all variables to be optimized based on the supervision
provided by the input images.

\paragraph{Notation.} Let $\mathcal{I} = \{\mathbf I_i\}_{1 \leq i \leq N}$ be the images captured at different timestamps and $\mathcal{P}$ be the aggregated LiDAR point clouds captured by a data collection platform, in our case a self-driving vehicle driving in the real world. Let $\{\mathbf M_i\}_{1 \leq i \leq N}$ be the foreground segmentation mask of $\{\mathbf I_i\}_{1 \leq i \leq N}$ obtained from an off-the-shelf algorithm \cite{zhang2021ners,kirillov2020pointrend}. Let $\mathcal{A} = \{\mathcal{D}, \mathcal{R}\}$ be the variables directly related to the appearance model (\ie, the material and the lighting).
We denote $ \Pi = \{\mathbf{K}^{\text{cam}}_i, \bm{\xi}^{\text{cam}}_i, \bm{\xi}^{\text{lidar}}\}$ as the intrinsics and the extrinsics of the sensors,
where $\bm{\xi} \in \mathfrak{se}$(3) are elements of the Lie algebra associated with $\text{SE}$(3).
We assume all cameras are pre-calibrated with known intrinsics.
Let $\psi: (\mathcal{M}, \mathcal{A}, \Pi) \rightarrow (\mathbf I_{\psi}, \mathbf M_{\psi})$ be the differentiable renderer where $\mathbf I_{\psi}$ and $\mathbf M_{\psi}$  denote the rendered RGB image and object mask.

\newcommand{\Ecolor}{E_\text{Color}}
\newcommand{\Emask}{E_\text{Mask}}
\newcommand{\Elidar}{E_\textsc{lidar}}

\paragraph{Overall energy.}
We design an energy function with complementary terms which measure
the geometry and appearance agreement between the observations and estimations ($E_\mathrm{data}$),
while regularizing the shape ($E_\mathrm{shape}$) and appearance ($E_\mathrm{app}$) to obey known priors:
\begin{align}
    \argmin_{\mathcal{M}, \Pi, \mathcal{A}} \{E_\mathrm{data}(\mathcal{M}, \Pi, \mathcal{A}; \mathcal{I}, \mathcal{P}) + \lambda_\mathrm{shape}  E_\mathrm{shape}(\mathcal{M}) + \lambda_\mathrm{app} E_\mathrm{app}(\mathcal{M}, \Pi, \mathcal{A}; \mathcal{I}, \mathcal{P})\}.
\label{eq:energy}
\end{align}
We now describe each energy term in more detail.

\paragraph{Data term.}
This energy encourages the estimated textured mesh to match the sensor data as much as possible.
Specifically, it consists of three components:
{\small
\begin{align}
\begin{split}
    E_\mathrm{data}(\mathcal{M}, \Pi, \mathcal{A}; \mathcal{I}, \mathcal{P}) = E_\mathrm{color}(\mathcal{M}, \Pi, \mathcal{A}; \mathcal{I}) &+ \lambda_\mathrm{mask}E_\mathrm{mask}(\mathcal{M}, \Pi; \mathcal{I}) + \lambda_\mathrm{lidar}E_\mathrm{lidar}(\mathcal{M}, \Pi; \mathcal{P}).
\end{split}
\end{align}
}%
$E_\mathrm{color}$, $E_\mathrm{mask}$ both enforce the rendering of the textured mesh to be close to the image observations. They, however, are complementary.
$E_\mathrm{color}$ encourages the appearance of the rendered image to match the RGB observation
and propagates the gradients to all variables including appearance variables $\mathcal{A}$,
while $E_\mathrm{mask}$ measures the mask difference and only depends on the shape of the mesh as well as the camera poses.
Following previous work~\citep{zhang2021ners,munkberg2021extracting}, we exploit smooth-$\ell_1$ distance to measure the difference in RGB space and squared $\ell_2$ for object masks:
{
\setlength{\belowdisplayskip}{0.5pt} \setlength{\belowdisplayshortskip}{0.5pt}
\setlength{\abovedisplayskip}{0.5pt} \setlength{\abovedisplayshortskip}{0.5pt}
\begin{align}
    E_\mathrm{color} = \frac{1}{N}\sum_{i}^{N} \mathrm{\bar{\ell}}_1\left(\mathbf{I}_\psi (\mathcal{M}, \mathbf{K}_i, \bm{\xi}_i, \mathcal{A}), \mathbf{I}_i\right) \quad \quad
    E_\mathrm{mask} = \frac{1}{N}\sum_{i}^{N} \left\lVert \mathbf{M}_\psi (\mathcal{M}, \mathbf{K}_i, \bm{\xi}_i) - \mathbf{M}_i \right \rVert^2_2,
\end{align}
}
with $\bar{\ell}_1$ the smooth-$\ell_1$ norm~\cite{huber1992robust} and $N$ the number of images available.

$E_\mathrm{lidar}$ encourages the geometry of our mesh to match the aggregated LiDAR point clouds.
Since minimizing point-to-surface distance is computationally expensive in practice, we adopt the popular Chamfer Distance (CD) to measure the similarity instead.
We randomly select $L$ points ($\mathcal{P}_\mathrm{s}$) %
from the current mesh and compute the asymmetric CD of $\mathcal{P}_\mathrm{s}$ with respect to the aggregated point cloud $\mathcal{P}$:
\begin{align}
E_\mathrm{lidar} = \mathrm{CD}(\mathcal{P},\mathcal{P}_\mathrm{s}) = \frac{1}{|\mathcal{P}|} \sum_{x \in \mathcal{P}} \alpha_x \min_{y \in \mathcal{P}_s} \lVert x - y \rVert^2_2,
\label{eq:lidar}
\end{align}
where $\alpha$ is an indicator function representing which LiDAR point is an outlier. We refer the reader to Sec.~\ref{sec:inference} for a discussion on how to estimate $\alpha$.

\paragraph{Shape term.}
This energy encourages the deformed mesh to be smooth and the faces of the mesh to be uniformly distributed among the surfaces (so that the appearance would be less likely distorted).
Specifically, $E_\mathrm{shape}$ is an addition of two shape regularizations~\cite{ravi2020pytorch3d,chen2021dibr++,zhang2021ners}: a normal consistency term $E_\mathrm{normal}(\mathbf V)$
 and an average edge length term {$E_\mathrm{edge}(\mathbf V)$}:
\begin{align}
E_\mathrm{normal}(\mathbf V) = \frac{1}{N_{\mathbf F}} \sum_{\mathbf f \in \mathbf F} \sum_{\mathbf f^\prime \in \mathcal{N}{(\mathbf f)}} || \mathbf n (\mathbf f) \cdot \mathbf n (\mathbf f^\prime)||^2_2 \quad
E_\mathrm{edge}(\mathbf V) = \frac{1}{N_{\mathbf E}} \sum_{\mathbf v \in \mathbf V} \sum_{\mathbf v^\prime \in \mathcal{N}{(\mathbf v)}} || \mathbf v - \mathbf v^\prime||^2_2.
\end{align}
Here, $\mathbf f \in \mathbf F$ and $\mathbf v \in \mathbf V$ refer to a single face and a single vertex. $\mathcal{N}(\mathbf f)$ and $\mathcal{N}(\mathbf v)$ denote the neighboring faces of $\mathbf f$ and the neighboring vertices of $\mathbf v$ respectively. $N_\mathbf{F}$ and $N_\mathbf{E}$ are the number of neighboring face pairs and edges.

\paragraph{Appearance term.}
This energy exploits the following facts: (1) the appearance of a vehicle will not change {abruptly in most cases}; instead, it varies in a smooth fashion; (2) neutral, white lighting dominates in the real world.
Following prior art in intrinsic decomposition \cite{land1971lightness,ma2018single},
we adopt a sparsity term to penalize frequent color changes on the diffuse $\mathbf{k}_d$ and specular $\mathbf{k}_s$ terms. We also add a regularizer to penalize the environment light in gray scale.
{
	\setlength{\belowdisplayskip}{0.5pt} \setlength{\belowdisplayshortskip}{0.5pt}
	\setlength{\abovedisplayskip}{0.5pt} \setlength{\abovedisplayshortskip}{0.5pt}
\begin{align}
E_\mathrm{app} = \lambda_\mathrm{mat} \left( \lVert \nabla \mathbf{k}_d \rVert_1 + \lVert \nabla \mathbf{k}_s \rVert_1 \right) + \lambda_\mathrm{light} \sum_{i = 1}^3 \lVert\mathbf{c}_i - \bar{\mathbf{c}}_i\rVert_1 ,
\end{align}
where $\nabla \mathbf{k}_d$ and $\nabla \mathbf{k}_s$ are the image gradients approximated by {Sobel-Feldman} operator~\cite{sobel19683x3}. $\mathbf{c}_i$ and $\bar{\mathbf{c}}_i$ are the light intensity values at R,G,B channels and the per-channel average intensities.
}

\subsection{Inference}
\label{sec:inference}

Our goal is to find the optimal mesh $\mathcal{M}^*$, appearance representation $\mathcal{A}^*$, and sensor poses $\Pi^*$ such that the total energy is minimized. We adopt the following strategy for better inference convergence.

\paragraph{Initialization.} A good initialization is required for non-convex optimization-based methods to achieve good performance.
If the energy model is initialized from a CAD asset that is very different from the observations, the mesh may not be able to fit the sensor data well.
To handle a wide variety of vehicles, we leverage the learned low-dimensional latent code $\mathbf{z}$ to estimate the initial coarse mesh $\mathcal{M}_\mathrm{init} = (\mathbf{V}_\mathrm{init}, \mathbf{F})$ , where $\mathbf{V}_\mathrm{init}$ is reconstructed from the optimized latent code $\mathbf{z}^*$ and
\vspace{0.03in}
{
	\setlength{\belowdisplayskip}{1pt} \setlength{\belowdisplayshortskip}{1pt}
	\setlength{\abovedisplayskip}{1pt} \setlength{\abovedisplayshortskip}{1pt}
\begin{align}
z^* = \argmin_z \lambda_\mathrm{mask}E_\mathrm{mask}(\mathcal{M}, \Pi; \mathcal{I}) + \lambda_\mathrm{lidar}E_\mathrm{lidar}(\mathcal{M}, \Pi; \mathcal{P}) + \lambda_\mathrm{shape} E_\mathrm{shape}(\mathbf{V}).
\end{align}
}%
We note that we focus on the geometry and only optimize the latent code. The
latent code is initialized from $\mathbf{0}$, and the sensor poses are obtained
from coarse calibration and fixed.
We optimize $\mathbf{z}$ using stochastic gradient descent with the
Adam optimizer \cite{kingma2014adam}. Please see the supp. material for additional
inference details.
Since the low-dimensional latent code already smooths out instance-specific details, the optimization is much more robust and less sensitive to initialization.

\paragraph{Efficient and robust inference.} Given the initialization $\mathcal {M}_\mathrm{init}$, we jointly optimize the vertices $\mathbf{V}$, appearance variables $\mathcal{A}$, and sensor poses $\Pi$ as described in Eq.~\eqref{eq:energy}.
To save computation, we uniformly sample $L$ points on the current mesh at each iteration to compute the LiDAR energy $E_\mathrm{lidar} = \mathrm{CD}(\mathcal{P}, \mathcal{P}_s)$, where $\mathcal{P}_s \sim \mathcal{M}$ and $|\mathcal{P}| = L$. To handle LiDAR outliers, we estimate the indicator function $\alpha$ by only calculating the Chamfer distance for the top $p$ percentage of point pairs with the smallest distance. %
We find it is sufficient to handle point outliers in most cases. For the sensor extrinsics, we optimize the 6D rotation representation~\cite{zhou2019continuity} and 3D translation.
It is a well-known issue that optimizing vertex positions with adverse gradient descent steps can cause self-intersections that is irrecoverable or even exacerbated in further steps~\cite{chen2021dibr++,nicolet2021large}.
Therefore, following~\citet{nicolet2021large}, we use preconditioned gradient descent steps to bias towards smooth solutions at each iteration, enabling faster convergence to higher-quality meshes with more fine-grained details. %

\section{Experiments}
\label{sec:result}

\subsection{Experimental Setup}
\label{sec:exp_setup}

\paragraph{Datasets.}
We evaluate our approach on two \emph{in-the-wild} datasets: Multi-View Marketplace Cars (MVMC)~\cite{zhang2021ners} and PandaVehicle~\cite{xiao2021pandaset}. MVMC consists of a diverse set of vehicles captured under various illumination conditions.
Each vehicle comes with around $10$ views and rough camera pose.
Following \cite{zhang2021ners}, we evaluate our model as well as the baselines on the selected 20 vehicles.
PandaVehicle is a dataset we derived from the PandaSet dataset~\cite{xiao2021pandaset}. It consists of 200+ vehicles, along with their corresponding multi-view images and aggregated LiDAR point clouds.
We manually inspect and select 10 high-quality vehicles for evaluation based on the accuracy of LiDAR-camera calibration. On average, each vehicle is captured from 24 camera views.

\paragraph{Baselines.} We compare our approach with three types of 3D reconstruction methods: (1) \textit{Neural radiance fields}: %
NeRF++~\cite{zhang2020nerf++}, and Instant-NGP~\cite{muller2022instant}.
(2) \textit{Implicit surface representations}:
NeRS~\cite{zhang2021ners}, NVDiffRec~\cite{munkberg2021extracting}, %
and NeuS~\cite{wang2021neus}.
(3) \textit{Explicit geometry-based approaches}: single-image view-warping (SI-ViewWarp) \cite{tulsiani2018layer}, which
warps the source image to the novel target view using depth estimated by LiDAR points; and multi-image view-warping (MI-ViewWarp), which blends multiple source images to the target view. We also compare against SAMP \cite{engelmann2017samp}, a CAD model mesh optimization approach which leverages SDF-aligned CAD models for joint pose and shape optimization.

\paragraph{Evaluation settings and metrics.}
{We evaluate reconstruction quality by performing camera and LiDAR simulation with the reconstructed vehicle.}
{For camera simulation (\ie, novel view synthesis (NVS))},
we adopt mean-square error (MSE), peak signal-to-noise ratio (PSNR), SSIM~\cite{wang2004image}, LPIPS~\cite{parmar2022aliased} and cleaned FID \cite{heusel2017gans}.
To focus on the reconstructed vehicle itself,
we only evaluate the foreground pixels using the {predicted} segmentation mask~\cite{kirillov2020pointrend}.
For LiDAR simulation, we place the reconstructed vehicle mesh in its original location and perform ray-casting to generate a simulated point cloud. We compare against held out real LiDAR points. We evaluate what fraction of real LiDAR points do not have a corresponding simulated point (\ie, coverage), the average per-ray $\ell_2$ error, Chamfer and Hausdorff distance.
See supp.\ for additional details.

\subsection{Reconstruction Quality}

\label{sec:results}

\begin{table}[]
	\centering
	\resizebox{0.8\textwidth}{!}{
	\begin{tabular}{@{}lcccccc@{}}
		\toprule
		\ \ Method        & MSE $\downarrow$ & PSNR $\uparrow$ & SSIM $\uparrow$ & LPIPS $\downarrow$ & $T$ (hour) & FPS \ \\ \midrule
		\midrule
		\ \ NeRF++~\mycite{zhang2020nerf++} & 0.0138 & 20.86 & 0.611 & 0.300 & 4.70 & 0.05 \\
		\ \ Instant-NGP~\mycite{muller2022instant} & 0.0095 & 21.68 & 0.641 & 0.319 & \underline{0.05} & 1.14 \\
		\midrule
		\ \ NeRS~\mycite{zhang2021ners} & 0.0176 & 18.49 & 0.562 & 0.265 & 1.37 & 3.23 \\
		\ \ NVDiffRec~\mycite{munkberg2021extracting} &
		0.0114 & 20.46 & 0.593 & 0.396 & 1.07 & \underline{51.2}* \\
		\ \ NeuS~\mycite{wang2021neus} & 0.0115 & 21.37 & 0.640 & 0.247 & 6.25 & 0.02 \\ \midrule
		\ \ SI-ViewWarp~\mycite{tulsiani2018layer} &  0.0233 & 17.51 & 0.514 & 0.371 &  $-$ & 1.67 \\
		\ \ SAMP~\mycite{engelmann2017samp} & 0.0144  & 19.52 & 0.628 & 0.283 & \underline{0.09} & \underline{71.4}* \\
		\ \ CADSim (ours) & \textbf{0.0087} & \textbf{21.72} & \textbf{0.674} & \textbf{0.220} & \underline{0.13} & \underline{49.6}*
		 \\
		\bottomrule
	\end{tabular}
}
	\vspace{0.03in}
	\caption{Evaluation of novel-view synthesis (left $\rightarrow$ front-left camera) on PandaSet. We report the average GPU hour $T$ for reconstruction per actor and rendering FPS on single RTX A5000. \underline{Underline} denotes fast reconstruction (\ie, $<$ 10 minutes) and real-time rendering (FPS $>$ 30).}
	\label{tab:eval_pandaset}
	\cuthalfcaptiondown
	\vspace{-0.13in}
\end{table}

\begin{figure}[t]
	\begin{center}
		\includegraphics[width=1.0\textwidth]{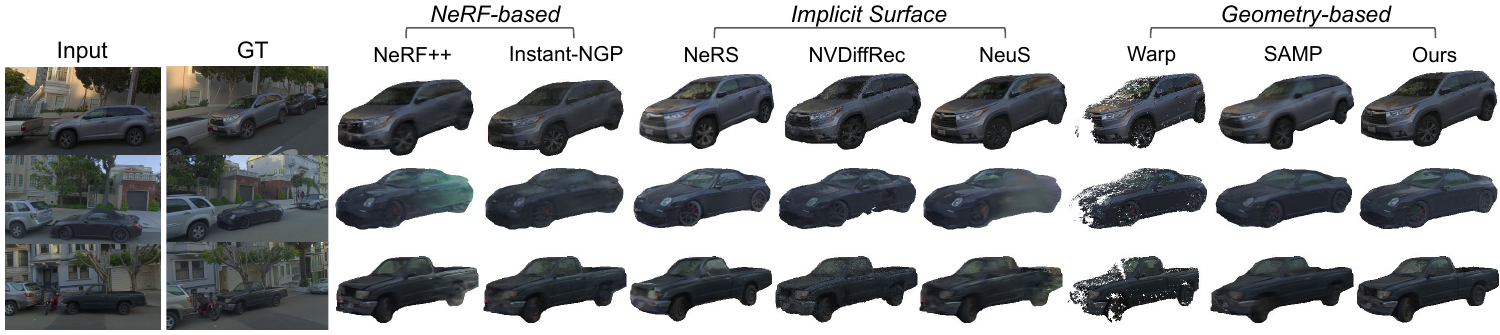}
	\end{center}
	\cuthalfcaptionup
	\caption{\textbf{Qualitative results on PandaVehicle for novel view synthesis}. Compared to existing reconstruction approaches, \model_name produces more robust and realistic results on large extrapolation.
	}
	\cuthalfcaptiondown
	\label{fig:qualitative_pandaset}
\end{figure}

\paragraph{Novel View Synthesis on {PandaVehicle} and {MVMC}.}
We report NVS results on {PandaVehicle} in Tab.~\ref{tab:eval_pandaset} and on {MVMC} in Tab.~\ref{tab:eval_mvmc}, along with reconstruction time ($T$) and
rendering speed (\ie, frame per second (FPS)).
For {PandaVehicle} we train each approach using images only from the left camera, and evaluate on front-left and front cameras.
For  {MVMC} we evaluate each approach via a cross-validation approach, where we hold out one image each time.
\model_name achieves better performance across all image quality metrics, especially LPIPS. LPIPS measures how neural nets perceive images. This suggests that our simulation results are perceived by the ML models to be more similar to real-world images (see Sec \ref{sec:applications} for more analysis).

\begin{wrapfigure}{R}{0.5\textwidth}
	\vspace{-0.15in}
	{\setlength{\tabcolsep}{3pt}
		\centering
		\resizebox{0.5\textwidth}{!}{
			\begin{tabular}{@{}lccccc@{}}
				\toprule
				\ \ Method        & MSE $\downarrow$ & PSNR $\uparrow$ & SSIM $\uparrow$ & LPIPS $\downarrow$ & FIDS $\downarrow$ \\ \midrule
				\ \ NeRS~\cite{zhang2021ners}          &  0.0254 & 16.5 & 0.720 & 0.172 & 60.9 \\
				\ \ CADSim &  \textbf{0.0188}  & \textbf{17.7} & \textbf{0.751} & \textbf{0.147}  & \textbf{48.1} \\ \bottomrule
			\end{tabular}
		}
		\vspace{-0.12in}
		\makeatletter\def\@captype{table}\makeatother%
		\caption{Evaluation of NVS on MVMC~\cite{zhang2021ners}.}
		\label{tab:eval_mvmc}
	}
	\vspace{0.05in}
	\centering
	{\setlength{\tabcolsep}{3pt}
		\resizebox{0.5\textwidth}{!}{
			\begin{tabular}{@{}lcccc@{}}
				\toprule
				\ Geometry & $L_2$ error $\downarrow$ & Hit rate $\uparrow$ & Chamfer$\downarrow$ &  Hausdorff$\downarrow$ \\ \midrule
				\ NeRS~\cite{zhang2021ners} & 0.171 & 94.3\% & 0.249 & 1.028 \\
				\ NVDiffRec~\cite{munkberg2021extracting} & 0.320 & \textbf{98.2}\% & 0.439 & 1.708 \\
				\ NeuS~\cite{wang2021neus} &
				0.367 & 90.3\% & 0.424 & 1.151
				\\
				\ SAMP~\cite{engelmann2017samp} & 0.158 & 94.8\% & 0.256 & 1.043 \\
				\rowcolor{grey} \ \model_name (ours) & \textbf{0.151}  & 96.3\% & \textbf{0.245} & \textbf{0.972} \\ \bottomrule
			\end{tabular}
		}
	}
	\vspace{-0.12in}
	\makeatletter\def\@captype{table}\makeatother%
	\caption{Comparison of LiDAR rendering metrics.}
	\vspace{-0.15in}
	\label{tab:compare_lidarsim}
\end{wrapfigure}

With CAD priors, we can reconstruct from much fewer views compared to NeRF-based or implicit-surface based-approaches.
Compared to geometry-based methods, our differentiable renderer and material model enable much higher photorealism.
Qualitative results on {PandaVehicle} in Fig.~\ref{fig:qualitative_pandaset} further confirm this.
NeRF-based approaches usually lead to visible artifacts with large view extrapolation and sparse input observations. %
Implicit surface approaches perform better by alleviating shape-radiance ambiguity \cite{zhang2020nerf++}, but still have artifacts due to lower quality geometries.
The other geometry-based approaches have better geometry, but have blurry appearance or missing pixels compared to \model_name.
To further demonstrate the value of our improved geometry,
we compare the NVS results using our reconstructed mesh against a NeRS mesh with various appearance representations in the supp.\ material (Tab. A1).
Additionally, neural rendering based approaches such as NeuS and NeRF++ are usually slow to train and render.
{While Instant-NGP \cite{muller2022instant} reduces training time significantly, it does not support real-time rendering. We report the rendering time for NVDiffRec, SAMP and CADSim (highlighted with *) using the differentiable renderer Nvdiffrast~\cite{Laine2020diffrast}. Faster rendering ($>$100 FPS) is expected with other modern graphics engines.

}

\vspace{-0.05in}
\paragraph{LiDAR simulation on PandaVehicle.}
We now evaluate the geometry performance of the reconstructed meshes for LiDAR simulation.
As shown in Tab.\ \ref{tab:compare_lidarsim}, our approach is better than or comparable to prior art across all metrics, suggesting that our meshes are more complete and more accurate than existing methods.
\mbox{NVDiffRec} has the highest hit rate, yet performs badly on the rest of the metrics. This is because their generated meshes are noisy and cover many LiDAR points.

\begin{figure}[t]
	\begin{center}
		\includegraphics[width=0.85\textwidth]{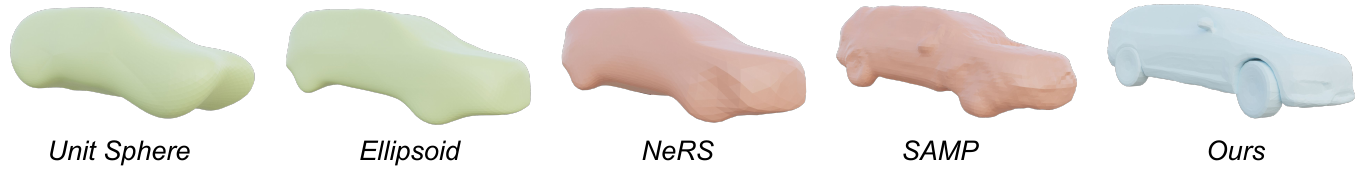}
	\end{center}
	\cuthalfcaptionup
	\vspace{-0.05in}
	\caption{CAD priors are crucial for achieving high-fidelity shape in non-convex optimization.}
	\vspace{-0.10in}
	\label{fig:cad_prior}
\end{figure}

\vspace{-0.05in}
\paragraph{Ablation study.}
To showcase the importance of CAD initalization, we initalize the mesh with the following alternatives:
a unit sphere, a rescaled ellipsoid, and geometries from either SAMP \cite{engelmann2017samp} or NeRS \cite{zhang2021ners}.
As shown in Fig.~\ref{fig:cad_prior}, exploiting CAD priors leads to faster convergence and better performance.
Our reconstruction is able to capture fine-grained details such as wheel size and rotation, rear-view mirrors, etc.
{We also evaluate our %
reflectance model choice~\cite{walter2007microfacet}  in the supp.\ material. %
Our physics-based material and lighting model performs the best and generalizes well to novel views.
}

\vspace{-0.05in}
\subsection{Realistic and Controllable Simulation}

\label{sec:applications}
We now showcase applying \model_name for accurate camera simulation to evaluate perception models, and to perform realistic vehicle insertion. \model_name also supports texture transfer naturally, allowing us to easily expand the asset library.
Please see additional details and results in the supp.\ material.

\vspace{-0.05in}
\paragraph{Downstream evaluation on camera simulation.}
To verify if \model_name helps reduce
domain gap for downstream perception tasks, we %
evaluate object detection and instance segmentation algorithms~\cite{he2017mask,kirillov2020pointrend} on simulated camera images at novel views.
In particular, we consider two setups: rendered and blended~\cite{chen2021geosim} and copy-pasted.
We compute the instance-level Intersection over Union (IoU) of the predicted bounding box and segmentation mask between simulated and real images.
This detection/segmentation agreement metric (the larger the
better) indicates how well the simulated data align with the real data with
respect to these two perception tasks and model architectures.
This metric can provide guidance on using simulation as one tool for evaluation.
As shown in Tab.~\ref{tab:downstream_evaluation},
using \model_name assets usually leads to the largest agreement with real images under different settings.

\setlength{\tabcolsep}{4.0pt}
\begin{table}[t]
	\centering
	\resizebox{0.9 \textwidth}{!}{
		\begin{tabular}{lccccccccccccc}
			\toprule
			& \multicolumn{5}{c}{\textit{Left camera $\rightarrow$ Front-left camera}} & &  \multicolumn{5}{c}{\textit{Left camera $\rightarrow$ Front camera}} \\ \cline{2-6}\cline{8-12}
			& \multicolumn{2}{c}{Blending~\cite{chen2021geosim}} & & \multicolumn{2}{c}{Copy-Paste} & & \multicolumn{2}{c}{Blending~\cite{chen2021geosim}} & & \multicolumn{2}{c}{Copy-Paste} \\
			& \multicolumn{1}{c}{Det. (IoU)} & \multicolumn{1}{c}{Segm. (IoU)} & & \multicolumn{1}{c}{Det. (IoU)} & \multicolumn{1}{c}{Segm. (IoU)} & & \multicolumn{1}{c}{Det. (IoU)} & \multicolumn{1}{c}{Segm. (IoU)} & & \multicolumn{1}{c}{Det. (IoU)} & \multicolumn{1}{c}{Segm. (IoU)} \\ \midrule
			Instant-NGP &  86.77 & 86.86 & & 80.57 & 80.81 & & 66.37 & 65.80 & & 41.65 & 40.13 \\
			NeuS & \underline{93.97} & \textbf{94.22} & &  92.82 & \underline{93.63} & & 66.26 & 65.09 &  & 66.01 & 64.66 \\
			SAMP & 90.39 &  89.58 & & 90.04 & 89.73 & & 82.82 & 79.78 & & \underline{83.32} & \textbf{81.79} \\
			\rowcolor{grey}\model_name (ours) & \textbf{94.15} & \underline{93.92} & & \textbf{93.71} & \textbf{93.72} & &  \textbf{83.99} & \textbf{80.94} & & \textbf{83.45} & \underline{81.42} \\
			\bottomrule
		\end{tabular}
	}
	\vspace{0.05in}
	\caption{Evaluation of downstream perception tasks (\ie, object detection, instance segmentation) on camera simulation, metric agreement with real data.
	Our assets lead to smaller domain gaps.
	}
	\vspace{-0.2in}
	\label{tab:downstream_evaluation}
\end{table}

\begin{figure}[htbp!]
	\vspace{-0.05in}
	\begin{center}
		\includegraphics[width=\textwidth]{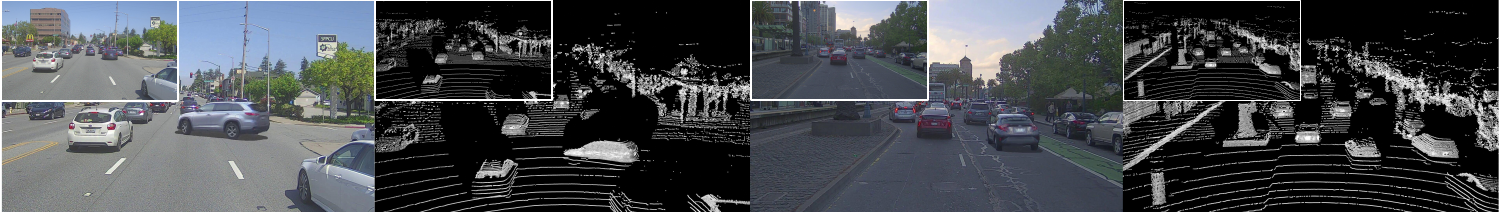}
	\end{center}
	\cuthalfcaptionup
	\caption{\textbf{Multi-sensor simulation on safety-critical scenarios with animated vehicle insertion.}
	The original camera and LiDAR observations captured in the real world are at the top-left corners. %
	}
	\vspace{-0.07in}
	\label{fig:actor_injection}
\end{figure}

\paragraph{Realistic animated vehicle insertion.}
Given the original scenario, sensor data, and an inserted actor trajectory,
we produce consistent multi-sensor simulation with the rendered asset.
In {Fig.~\ref{fig:actor_injection}}, we generate safety-critical scenarios by inserting a new actor turning into our lane
(left half)
and an actor slowing down in the adjacent lane, preventing the SDV from lane-changing (right half). The LiDAR simulation is conducted on reconstructed meshes enriched with intensity values (see supp. A2). The rendered asset is seamlessly blended into the original scenario and has physically realistic wheel rotation and movement.
This enables diverse scenario generation for end-to-end autonomy testing.

\section{Limitations and Conclusion}

\model_name's
main assumption is that it requires CAD models for the object class of interest.
We note that CAD models are readily available for most object classes, and that our approach only requires encoding semantic priors for a single CAD asset, as our energy-model optimization allows for transfer of these priors to other assets of the same class. See supp. for non-vehicle classes.
Additionally, our approach relies on either
good segmentation masks or LiDAR points to reconstruct accurate shapes.

In this paper, we proposed to leverage in-the-wild camera and LiDAR data to reconstruct objects such as vehicles.
Towards this goal, we designed \model_name, which leverages geometry and semantic cues from CAD models with differentiable rendering, to generate meshes with high quality geometry and appearance. %
These cues also enable our approach to generate \emph{articulated} and \emph{editable} meshes, enabling endless creation of new shapes, textures, and animations for simulation.
We demonstrated that integrating our meshes into a camera simulation system can more effectively evaluate perception algorithms than existing reconstruction methods, reducing the simulation-to-real domain gap.

\section*{Acknowledgement}

We sincerely thank the anonymous reviewers for their insightful suggestions. We would like to thank Chris Zhang and Yuwen Xiong for their feedback on the early draft and final proofreading. We also thank the Waabi team for their valuable assistance and support.

\bibliography{corl_2022}  %
\clearpage

\appendix
\section*{Appendix}
\renewcommand{\thetable}{A\arabic{table}}
\renewcommand{\thefigure}{A\arabic{figure}}

\section{CADSim Implementation Details}

\subsection{Learning a shared representation}
\label{sec:cad_lib}
\begin{figure}[t]
	\begin{center}
		\includegraphics[width=0.98\textwidth]{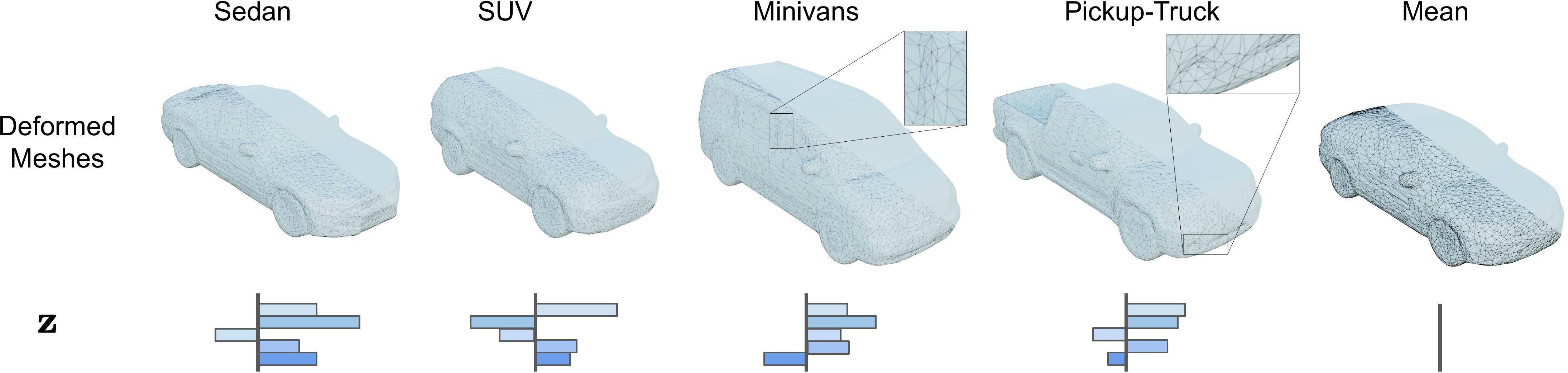}
	\end{center}
	\cuthalfcaptionup
	\caption{\textbf{Collect watertight and vertex-aligned vehicle meshes from CAD models}. Given a collection of CAD vehicles, we first use a robust algorithm~\citep{huang2018robust,huang2020manifoldplus} to obtain watertight manifolds. Then, we pick one simplified mesh as the source mesh and deform it to the other meshes by minimizing the shape energy. Finally, we build the PCA basis on deformed meshes with latent codes stored for each object.}
	\cuthalfcaptiondown
	\label{fig:car_pca}
\end{figure}

As discussed in Sec 3, we create a shared representation space over a set of CAD models to handle a wide variety of vehicle shapes during optimization.
Specifically, we apply principal component analysis (PCA) on the vertex coordinates {of the CAD models to obtain a shared low-dimensional code $\mathbf z$}.
Formally, we have
\begin{align}
\begin{split}
\mathbf z &= \mathbf W^\top \left( \mathbf V - \boldsymbol{\mu} \right), \\
\widehat{\mathcal{M}} &= \{\widehat{\mathbf V}, F\}, \quad \mathrm{where} \ \
\widehat{\mathbf V} = \mathbf W \mathbf z + \boldsymbol{\mu}.
\end{split}
\end{align}
$\boldsymbol \mu \in \mathbb R^{|V| \times 3}$ is the mean vertices of the
meshes, and $\mathbf W$ is the top $K$ principle components. $ \mathbf z$, $\widehat{\mathcal{M}}$ and $\widehat{\mathbf V}$ are the latent code, reconstructed mesh and vertices respectively. We note that since all the CAD models are parameterized with Eq.~(1) from the main paper, the reconstructed mesh $\widehat{\mathcal{M}}$ by nature consists of parts and supports wheel articulation. As shown in Fig.~\ref{fig:car_pca} (first row), the deformed meshes
maintain important geometry details of vehicles such as rear-view mirrors and car grilles.

\paragraph{CAD library alignment.} In order to learn a low dimension code over a variety of vehicles, we must align the templates from different CAD models and establish a one-to-one dense correspondence among the vertices.
The original CAD models are unaligned, as they have a varying number of vertices and the vertex ordering differs across models.
Specifically, {we select a single template mesh $\mathcal M_\mathrm{src}$} as the source mesh and deform its vertices $\mathbf V$ such that it fits other meshes well. We exploit the vertices of the simplified target mesh (denoted as $\mathcal{P}_\mathrm{cad}$) and minimize the following energy:
\begin{align}
E_\mathrm{align}(\mathbf V, {\mathcal{P}_\mathrm{cad}}) &= E_\mathrm{chamfer}(\mathbf V, \mathcal{P}_\mathrm{cad}) + \lambda_\mathrm{shape} \cdot E_\mathrm{shape}(\mathcal{M_\mathrm{src}}).
\end{align}
Here, $E_\mathrm{chamfer}$ refers to the asymmetric Chamfer distance, and $E_\mathrm{shape}$ is the same as described in Sec. 3.2.
We note that this is an offline procedure separate from the energy minimization in Sec 3.3.

\subsection{Appearance Representation}
\label{sec:appearance}
In addition to geometry, our mesh representation must accurately capture how light interacts with its surface so that we can realistically reproduce sensor observations. Towards this goal, we parameterize the mesh appearance using a physically-based appearance representation.
Specifically, we represent appearance using a micro-facet BRDF model with the differentiable split sum environment lighting~\cite{karis2013real} proposed by~\citet{munkberg2021extracting}.
Since the topology of the mesh is fixed, we can build a one-to-one mapping relationship between each point on the mesh surface and each point in the 2D $(u, v)$ space.
We use a Physics-Based Rendering (PBR) material model from Disney~\cite{mcauley2012practical}, which contains a diffuse lobe $\mathbf k_d$ with an isotropic, specular GGX lobe~\cite{walter2007microfacet}.
Following the standard convention, we store them together with normals in three
texture images $\mathbf k_d$, $\mathbf k_\text{orm}$ and $\mathbf k_n$.
$\mathbf k_\text{orm} = (o, r, m)$ where $o$ is left unused, $r$ is the roughness value and $m$ is the metalness factor that interpolates between plastic and metallic appearance by computing a specular highlight color: $\mathbf k_s = (1 - m) \cdot 0.04 + m \cdot \mathbf k_d$.
For the lighting model, we use a differentiable version of the split sum shading model introduced by~\citet{munkberg2021extracting}. Specifically, we optimize a cube map ($6 \times 512 \times 512$), where the base level represents the pre-integrated lighting for the lowest supported roughness value and each smaller mip-level is reconstructed using~\cite{hill2020physically}. We refer the readers to~\cite{munkberg2021extracting} for more details.

\paragraph{Intensity retrieval.} For LiDAR simulation, the optimized meshes are enriched with per-vetex intensity by retrieving the top 10 closest points for aggregated point cloud and taking the average intensity value. Empirically we find the performance is similar to direct energy optimization on the vertex intensity.

\subsection{Inference Details}
\label{sec:details}
Since all operations are differentiable, one straightforward way to conduct inference is to directly minimize the full energy with gradient-based methods. Unfortunately, due to the highly non-convex structure of the energy model as well as the noise in the observations, such an approach will often lead to sub-optimal solutions. We thus adopt the following curriculum strategy.

First, we optimize the latent code $\mathbf z$ initialized from $\mathbf 0$ with Adam optimizer (learning rate $\text{3e-2}$) for 200 iterations (Eq. 8 in the main paper), while keeping the other variables fixed.
Given the initialization $\mathcal {M}_\mathrm{init}$, we jointly optimize the vertices $\mathbf{V}$, appearance variables $\mathcal{A}$, and sensor poses $\Pi$ as described in Eq. (3).
At the beginning of the optimization (first 500 iterations), we do not add $E_\mathrm{color}$ term so that the model will focus on geometry (\eg, $E_\mathrm{mask}$, $E_\mathrm{lidar}$, etc). The hyperparameters are set as $\lambda_\mathrm{mask} = \lambda_\mathrm{lidar} = 0.5, \lambda_\mathrm{shape} = 0.1$. To enforce the  learned geometry for vehicles to be symmetric, we flip the vertices of optimized mesh along the symmetry axis and add a symmetric Chamfer distance ($\lambda_\mathrm{sym} = 0.5$) between the original and flipped meshes. We use AdamUniform~\cite{nicolet2021large} (learning rate 3e-2) to optimize the vertices and standard Adam (learning rate 1e-4) to optimize the intrinsics and the extrinsics of the sensors $\Pi$.
Then in the next 500 iterations, we add $E_\mathrm{color}$ term to optimize the appearance variables $\mathcal{A}$ as well as refine the sensor poses and geometry through RGB information. We set the coefficients for appearance energy term as follows: $\lambda_\mathrm{app} = 1, \lambda_\mathrm{mat} = \text{1e-9}, \lambda_\mathrm{light} = \text{1e-2}$.
We use the Adam optimizer \cite{kingma2014adam} with a learning rate scheduler with an exponential falloff from 0.03 to 0.01 over 500 iterations. Empirically we find this simple weighting strategy very effective.

\section{Implementation Details for Baselines}
\label{sec:details_baseline}
We compare our approach with a wide range of state-of-the-art 3D reconstruction methods, which can be roughly divided into: (1) \textit{Neural radiance fields}: %
NeRF++~\cite{zhang2020nerf++}, and Instant-NGP~\cite{muller2022instant}.
(2) \textit{Implicit surface representations}:
NeRS~\cite{zhang2021ners}, NVDiffRec~\cite{munkberg2021extracting}, %
and NeuS~\cite{wang2021neus}.
(3) \textit{Explicit geometry-based approaches}: single-image view-warping (SI-ViewWarp) \cite{tulsiani2018layer}, which
warps the source image to the novel target view using depth estimated by LiDAR points, and multi-image view-warping (MI-ViewWarp), which blends multiple source images to the target view. We also compare against SAMP \cite{engelmann2017samp}, a CAD model mesh optimization approach which leverages SDF-aligned CAD models for joint pose and shape optimization. In the following, we provide the overview and implementation details of these reconstruction baselines.

\subsection{NeRF-based Approaches}
\paragraph{NeRF++~\cite{zhang2020nerf++}.} NeRF++ introduced an inverse sphere parameterization to extending NeRF to large-scale and unbounded 3D scene. To train NeRF++ properly on PandaVehicle, we first scale the scene to normalize all cameras' position within a unit sphere. Then we adopt the same hyperparameters from the official code repository\footnote{\url{https://github.com/Kai-46/nerfplusplus}} except that we train for 100k iterations because PandaVehicle has fewer views and the model converges with fewer iterations.
\paragraph{Instant-NGP~\cite{muller2022instant}.} Instant-NGP achieved the state-of-the-art performance by introducing the efficient hash encoding and fully fused MLPs. In our experiments, we properly scale the objects to be reconstructed in the unit cube and set \texttt{aabb\_scale} as 4 to handle the background visible outside the unit cube. We tuned \texttt{aabb\_scale} to be 1, 2, 4, 8 and 16 and find 4 leads to the best performance. The model is trained for 5k iterations and converges on the training views. During inference, we render with 4 samples per pixel for better results (effectively doing 4x superresolution for anti aliasing).

\subsection{Implicit Surface Representations.}
\paragraph{NeRS~\cite{zhang2021ners}.} Zhang et al.\ proposed a neural surface representation combined with differentiable rendering to generalize with sparse in the wild data. We followed the official  NeRS\footnote{\url{https://github.com/jasonyzhang/ners}} implementation. On MVMC, we use the official evaluation code and focus on the ``fixed optimized camera'' setting. For PandaVehicle, we initialized the cuboid template with the assets' coarse 3D dimensions and set the level of unit ico-sphere as 6. We employ a three-stage training process: sequentially optimizing the shape, texture, and illumination parameters.
To ensure better visual quality and semantic metrics, we increased the weights of the chamfer loss and perceptual loss to 0.04 and 1.0, respectively. We also removed off-screen loss as not all input views contain the complete vehicle shapes. Moreover, we increased the training iterations on the three stages to $3k, 12k, 3k$ since more input views are provided in PandaVehicle and it takes longer to converge. We also applied symmetry constraints to the deformed textured meshes along the heading axis.

\paragraph{NVDiffRec~\cite{munkberg2021extracting}.} Munkberg et al.\ proposed an efficient differentiable rendering-based reconstruction approach that combines differentiable marching tetrahedrons and split-sum environment lighting. It achieves the state-of-the-art performance on a wide variety of synthetic datasets with dense camera views. We follow the official code implementation\footnote{\url{https://github.com/NVlabs/nvdiffrec}}. We set the tetrahedron grid resolution as $64$ and the mesh scale as $5.0$ (real vehicle scale). The model is trained for 5k iterations (batch size 8) with a learning rate exponentially decayed from 0.03 to 0.003.

\paragraph{NeuS~\cite{wang2021neus}.} We follow the official code repository\footnote{\url{https://github.com/Totoro97/NeuS}} and train each asset for 200k iterations.
We perform scene normalization to make the asset's region of interest fall inside a unit sphere, and model the background by NeRF++~\cite{zhang2020nerf++}.
To train the NeuS with LiDAR supervision, we render the LiDAR depth map, and use it to supervise the volume rendered depth similar to DS-NeRF~\cite{deng2021depth}.

\subsection{Geometry-based Approaches.}

\paragraph{SI-ViewWarp~\cite{chen2021geosim}.}
For warping based methods, we follow the implementation in GeoSim \cite{chen2021geosim}. Give an asset mesh, we first render the mesh at the target viewpoint to generate the target depth map. Then, we un-project the target depth map to source images and get the corresponding pixel color using the inverse warping operations. We compare un-projected depth in source view and asset-rendering depth in source view to filter invisible region. This approach doesn't need training for rendering. To get the best rendering results, we warp using all source images as candidates, and heuristically pick the one with minimum unseen region in the target view.

\paragraph{MI-ViewWarp.} Since there are often unobserved regions at the target view when warping from a single source view, we further extend SI-ViewWarp to warp from multiple source images progressively (MI-ViewWarp). Specifically, we sort the source images in an increasing order according to the distance of the viewpoints (or camera matrices) between source and target images. Then we iterate on the sorted source images, conduct SI-ViewWarp and fill in the pixels progressively only if not occupied. We find this heuristic warping strategy works
well in practice as it takes the ``confidence" (according to distance) of each source image into account. Compared with simple averaging of all warped single images, our strategy produces non-blurry results and better metrics.

\paragraph{SAMP~\cite{engelmann2017samp}.} Given the processed CAD library where each mesh is watertight and simplified, we compute volumetric SDFs for each vehicle in metric space (volume dimension $100 \times 100 \times 100$). Following~\cite{engelmann2017samp,duggal2022secrets}, we apply PCA on the SDF volumes and set the embedding dimension as 25. During inference, we jointly optimize the shape latent code, a scaling factor on the SDF (handle diverse shapes) and relative vehicle pose (rotation, translation) to fit the LiDAR points. We adopt the $\mathcal L_1$ loss on the SDF difference and a total variation loss on the scale factor to penalize abrupt local SDF changes. The weights of data and regularization terms are 1 and 0.1. We use the Adam optimizer with a learning rate of 0.01. The mesh is extracted from the SDF volume via marching cubes \cite{lorensen1987marching}. Given the optimized shape, we then use a similar CADSim-like energy minimization procedure to optimize a 2D UV texture image using a differentiable renderer. Compared with CADSim, optimized SAMP geometries usually have fewer
fine-grained details and have worse image alignments.

\section{Dataset and Metric Details}
\label{sec:dataset_metrics}
\subsection{PandaVehicle Details}
\label{sec:panda_vehicle}

We derive PandaVehicle from the PandaSet~\cite{xiao2021pandaset} dataset. PandaSet is a dataset captured by a self-driving vehicle platform equipped with six cameras (left, front left, front, front right, right and back cameras) and two LiDARs (a top $360^\circ$ mechanical spinning LiDAR and a forward-facing LiDAR). In PandaVehicle, we select vehicles that are observed by the main top Pandar64 sensor, as well as the left, front left and front cameras. %
We observe that there is poorer calibration for the front right, right and back cameras and therefore do not use them for quantitative evaluation.
We train or reconstruct vehicle meshes with observations from the top LiDAR and the left camera images only. In the novel view synthesis task, we test by rendering the mesh and comparing with images from the front left and front cameras. We generate all semantic segmentation masks with PointRend~\cite{kirillov2020pointrend}.

We selected 10 vehicles that have high-quality camera-LiDAR alignment for evaluation. Fig.~\ref{fig:pandaset} shows the aggregated LiDAR points and images used for one of the selected vehicles. Table~\ref{tab:pandaset-actors} includes detailed information for all 10 selected vehicles.

\subsection{LiDAR Rendering Metrics}
\label{sec:lidar_metrics}
We provide additional details on how we compute LiDAR rendering metrics used in the main paper Table 3.
Given the aggregated point clouds $\mathcal{P}$, we apply voxel downsampling with a resolution of 5cm to obtain the input point clouds $\mathcal{P}_\mathrm{input}$ for reconstruction. Then the held-out
real LiDAR points can be written as $\mathcal{P}_\mathrm{held} = \mathcal{P} \setminus \mathcal{P}_\mathrm{input}$. We evaluate the average per-ray $\ell_2$ error and what fraction of real held-out LiDAR points have a corresponding simulated point (\ie, Hit rate).
Lastly, we place the reconstructed vehicle mesh in its original location and perform ray-casting to generate a simulated point cloud and report the Chamfer and Hausdorff distance with original aggregated point clouds $\mathcal{P}$.

\section{Additional Experiments and Analysis}
In this section, we show additional experiments on MVMC and PandaVehicle, including geometry comparison with NeRS on MVMC, NVS with larger extrapolation, and ablation studies on LiDAR supervision, mesh initialization and appearance representation. We also show additional downstream perception results.

\subsection{Ablation Experiments on MVMC and PandaVehicle}
\label{sec:additional_exp}

\begin{table}[htbp!]
	\centering
	\resizebox{0.8\textwidth}{!}{
		\begin{tabular}{@{}lcccccc@{}}
			\toprule
			\ \ Method                            & Geometry & MSE $\downarrow$ & PSNR $\uparrow$ & SSIM $\uparrow$ & LPIPS $\downarrow$ & FIDS $\downarrow$ \\ \midrule
			\ \ \multirow{2}{*}{SI-ViewWarp~\cite{tulsiani2018layer}} & NeRS     & 0.0683 &  12.3 & 0.659 & 0.288 &   93.5     \\
			& Ours     & \textbf{0.0587} &  \textbf{12.9} & \textbf{0.674} & \textbf{0.257}  & \textbf{79.1}  \\ \midrule
			\ \ \multirow{2}{*}{MI-ViewWarp}  & NeRS     &  0.0430  & 14.4 & 0.665 & 0.225  & 67.2  \\
			& Ours     & \textbf{0.0311} & \textbf{15.6} & \textbf{0.687} & \textbf{0.185} &  \textbf{51.4}   \\ \midrule
			\ \ \multirow{2}{*}{UV Map Opt.}           & NeRS     & 0.0371 & 15.1 & 0.690 & 0.215 &   74.6    \\
			& Ours     &  \textbf{0.0215} &  \textbf{17.1} & \textbf{0.725} & \textbf{0.160}  & \textbf{51.6}    \\ \bottomrule
		\end{tabular}
	}
	\vspace{0.03in}
	\caption{Comparison of geometry quality with NeRS~\cite{zhang2021ners} on MVMC.}
	\vspace{-0.1in}
	\label{tab:geo_mvmc}
\end{table}

\begin{table}[htbp!]
	\centering
	\resizebox{0.8\textwidth}{!}{
		\begin{tabular}{@{}lcccc@{}}
			\toprule
			\ \ Method        & MSE $\downarrow$ & PSNR $\uparrow$ & SSIM $\uparrow$ & LPIPS $\downarrow$ \ \\ \midrule
			\ \ Instant-NGP~\mycite{muller2022instant} & 0.0302 & 15.68 &  0.417 &  0.479
			\\
			\ \ NeuS~\mycite{wang2021neus} & 0.0280 & 16.17 & 0.403 & 0.439 \\
			\ \ SAMP~\mycite{engelmann2017samp} & 0.0258  & 16.42 & 0.401 & 0.362 \\
			\ \ CADSim (ours) & \textbf{0.0218} & \textbf{16.87} & \textbf{0.424} & \textbf{0.334} \\
			\bottomrule
		\end{tabular}
	}
	\vspace{0.03in}
	\caption{Evaluation of novel-view synthesis on extreme views (left $\rightarrow$ front camera) on PandaSet.
	}
	\label{tab:extreme_nvs}
\end{table}

\paragraph{Geometry Comparison on MVMC.} To further demonstrate CADSim is able to reconstruct higher-quality geometries,
we report the NVS results comparing our reconstructed mesh compared to NeRS \cite{zhang2021ners} mesh on various appearance representations. We follow the same evaluation setting as Table 1 in the main paper.
We only change how the texture for the mesh is created, while keeping the optimized geometry fixed.
As shown in Table~\ref{tab:geo_mvmc}, our geometry consistently leads to better NVS performance due to better alignment with images and improved optimization
with CAD priors.

\begin{figure}[t]
	\begin{center}
		\includegraphics[width=0.98\textwidth]{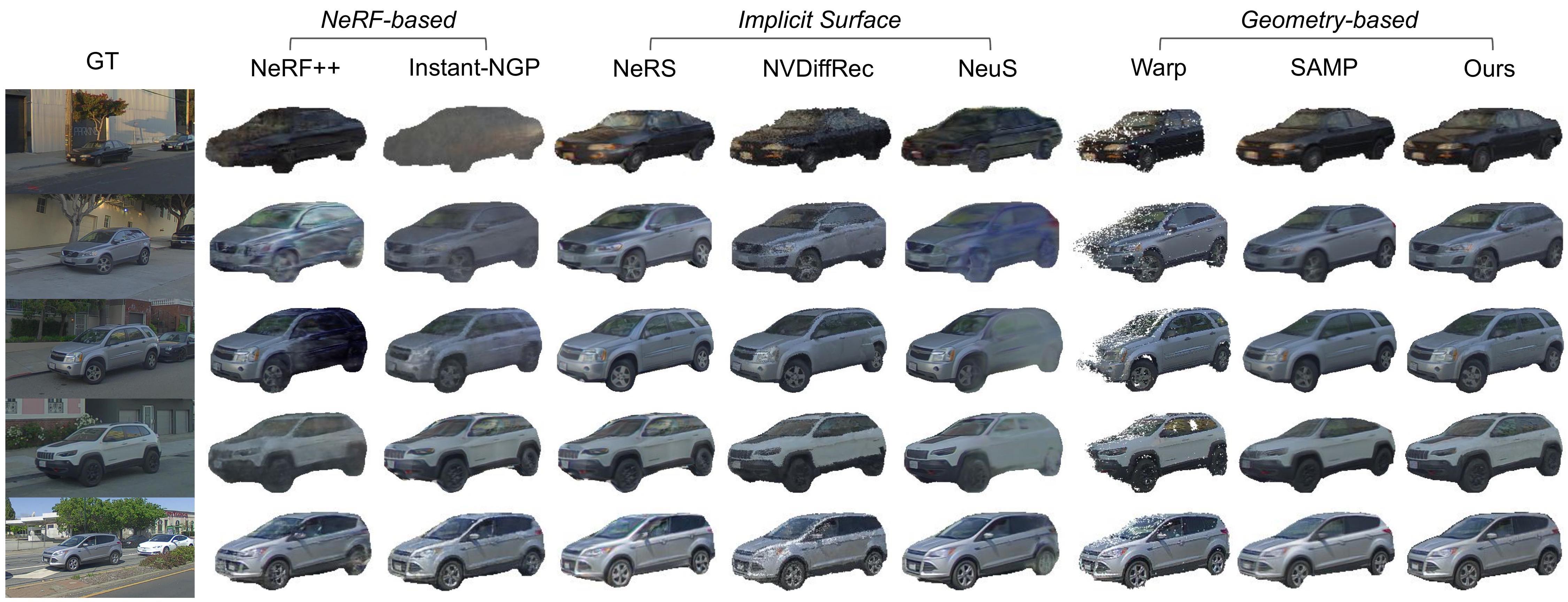}
	\end{center}
	\cuthalfcaptionup
	\caption{
		\textbf{Additional qualitative results on PandaVehicle for novel view synthesis.} Compared to existing reconstruction approaches, CADSim produces more robust and realistic results on large extrapolation.
	}
	\cuthalfcaptiondown
	\label{fig:qualitative_comp}
\end{figure}

\paragraph{Additional Qualitative Results.} We provide additional qualitative results on PandaVehicle in Figure~\ref{fig:qualitative_comp}. CADSim is able to outperform all baselines consistently on large extrapolation setting. This is because we leverage CAD priors during shape and appearance optimization and produce higher quality geometry. In contrast, NeRF-based and implicit surface approaches usually suffer from obvious artifacts due to shape-radiance ambiguity~\cite{zhang2020nerf++}. The other geometry-based approaches either relies on non-manifold surfel meshes or coarse volumetric SDFs thus leading to missing pixels, blurry results and large image-geometry misalignment.

\begin{figure}[htbp!]
	\begin{center}
		\includegraphics[width=\textwidth]{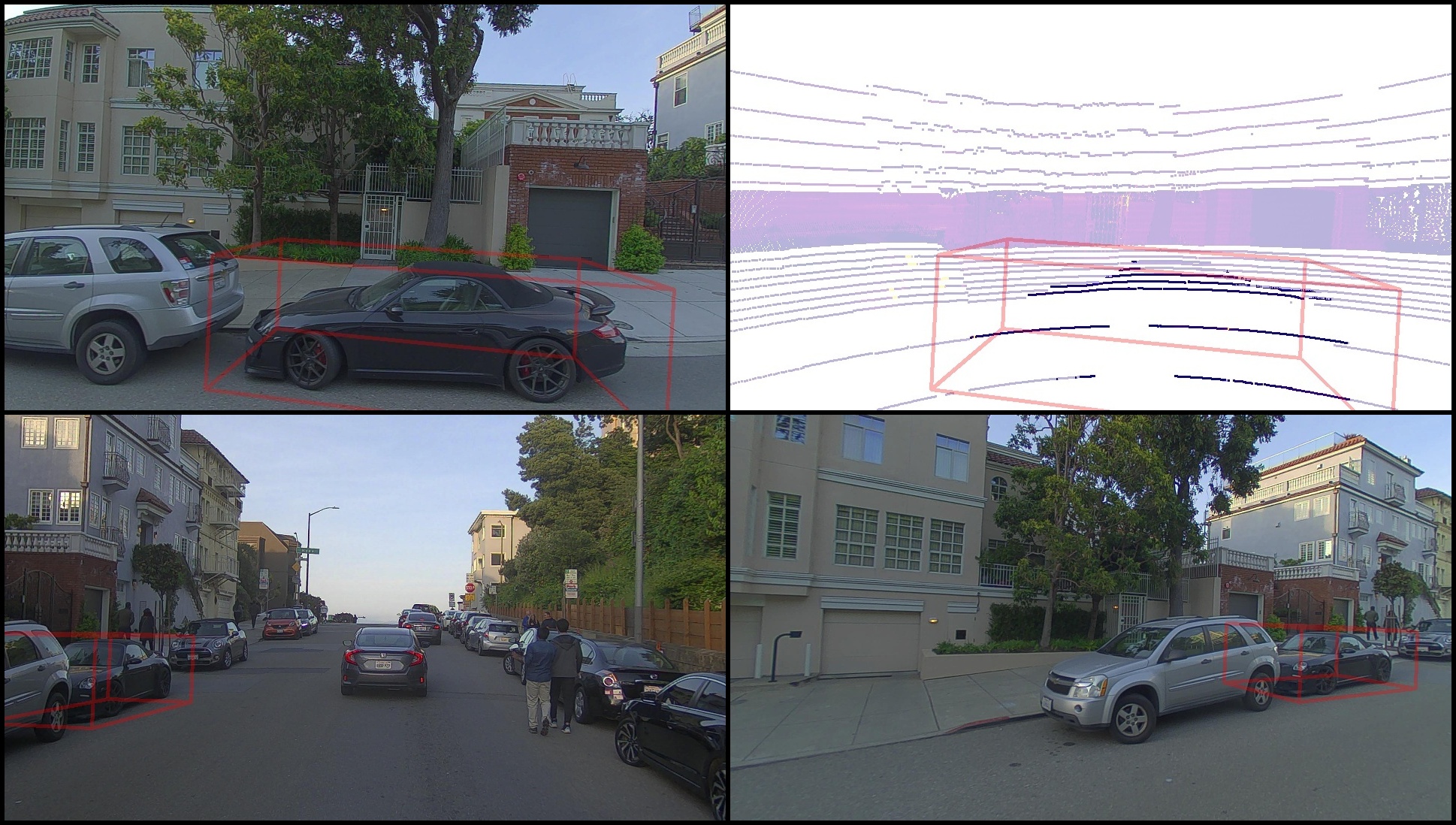}
	\end{center}
	\cuthalfcaptionup
	\caption{Illustration of vehicle \texttt{526e2f5e-e294-415c-aad6-578d27921465} (in the red bounding box) in Pandaset sequence \texttt{030}. (Top) Image and %
		LiDAR points from the left camera view used during mesh reconstruction. (Bottom) Front camera and front left camera views used during testing for the novel view synthesis task.}
	\label{fig:pandaset}
\end{figure}

\paragraph{NVS for Extreme Views on PandaVehicle.} To test the robustness of CADSim in the self-driving context, we conduct a more challenging NVS task: left camera $\rightarrow$ front camera.
Fig.~\ref{fig:pandaset} (left) visualizes the significant change in viewpoint.
Since there is no
overlap in the field-of-view between left and front cameras on Pandaset\footnote{https://scale.com/open-datasets/pandaset}, it requires better geometry and appearance for large extrapolation. We pick the baselines with the best performance on each category in Sec~\ref{sec:details_baseline}. As shown in Table~\ref{tab:extreme_nvs}, our approach results in the best performance, especially on LPIPS which measures how well the machine learning models perceive the simulated images.

\paragraph{Investigation of LiDAR Supervision.} Compared to previous reconstruction approaches, CADSim can leverage the LiDAR points that are usually accessible in the self-driving applications. While it is not our focus to extend existing baselines to leverage LiDAR properly, we choose one of the best performing baselines, NeuS, and use the rendered LiDAR depth map to supervise the volume rendered depth, similar to DS-NeRF~\cite{deng2021depth}. Moreover, we also remove the LiDAR branch $E_\mathrm{lidar}$ of CADSim for comparison. As shown in Table~\ref{tab:eval_pandaset_depth} and Table~\ref{tab:depth_lidar}, adding depth supervision for NeuS helps improve the NVS performance (especially on LPIPS) and LiDAR rendering results by reducing the ambiguity caused by sparse views. We note that although the LiDAR rendering metrics have improved, the geometry obtained by NeuS is still not complete.
This may lead to a dramatic performance decrease when we place actors in completely new
locations and viewpoints for simulation. In contrast, CADSim is less dependent on LiDAR supervision and our assets are always complete with the help of CAD shape priors.

\begin{table}[htbp!]
	\centering
	\resizebox{0.7\textwidth}{!}{
		\begin{tabular}{@{}lcccc@{}}
			\toprule
			\ \ Method        & MSE $\downarrow$ & PSNR $\uparrow$ & SSIM $\uparrow$ & LPIPS \ \\ \midrule
			\midrule
			\ \ NeuS~\mycite{wang2021neus} & 0.0115 & 21.37 & 0.640 & 0.247  \\
			\ \ NeuS + LiDAR & 0.0103 & 21.42 & 0.649 & 0.228 \\ \midrule
			\ \ CADSim - LiDAR & 0.0096 & 21.59
			& 0.665 & 0.231 \\
			\ \ CADSim (ours) & \textbf{0.0087} & \textbf{21.72} & \textbf{0.674} & \textbf{0.220}
			\\
			\bottomrule
		\end{tabular}
	}
	\vspace{0.03in}
	\caption{Evaluation of novel-view synthesis (left $\rightarrow$ front-left camera) on PandaSet. }
	\label{tab:eval_pandaset_depth}
\end{table}

\begin{table}[htbp!]
	\centering
	\vspace{-0.15in}
	\resizebox{0.75\textwidth}{!}{
		\begin{tabular}{@{}lcccc@{}}
			\toprule
			\ Geometry & $L_2$ error $\downarrow$ & Hit rate $\uparrow$ & Chamfer$\downarrow$ &  Hausdorff$\downarrow$ \\ \midrule
			\ NeuS~\cite{wang2021neus} &
			0.367 & 90.3\% & 0.424 & 1.151
			\\
			\ NeuS + LiDAR & 0.191 & 95.3\% & 0.261 & 1.017 \\ \midrule
			\ \model_name - LiDAR & 0.155  & 95.2\% & 0.249 & 0.988 \\
			\ \model_name (ours) & \textbf{0.151}  & \textbf{96.3\%} & \textbf{0.245} & \textbf{0.972} \\ \bottomrule
		\end{tabular}
	}
	\vspace{0.03in}
	\caption{Comparison of LiDAR rendering metrics.}
	\label{tab:depth_lidar}
\end{table}

\paragraph{Geometry and Reflectance Model Ablation Study.}
We rigorously evaluate that our method's use of a CAD model improves performance in Table~\ref{tab:ablation_init}, where
we replace the proposed CAD initialization with the following alternatives:
a unit sphere, a rescaled ellipsoid, and geometries from either SAMP \cite{engelmann2017samp} or NeRS \cite{zhang2021ners}.
Our CAD mesh initialization has the highest performance.
We also evaluate our %
reflectance model choice~\cite{walter2007microfacet} in Table~\ref{tab:ablation_tex}. Our physics-based material and lighting model performs the best and generalizes well to novel views.

\begin{figure}[htbp!]
	\begin{minipage}{\textwidth}
		\begin{minipage}[c]{0.47\textwidth}
			\begin{table}[H]
				\resizebox{\linewidth}{!}{
					\centering
					\begin{tabular}{@{}lcccc@{}}
						\toprule
						\ \ Initialization        & PSNR $\uparrow$ & SSIM $\uparrow$ & LPIPS $\downarrow$ \\ \midrule
						\ \ Sphere & 20.61 & 0.631 & 0.243 \\
						\ \ Ellipsoid & 21.10 & 0.653 & 0.233  \\
						\ \ SAMP~\cite{engelmann2017samp} & 21.32 & 0.667 & 0.230 \\
						\ \ NeRS~\cite{zhang2021ners} &  21.58 & 0.665 & 0.235 \\
						\ \ CAD (ours) & \textbf{21.72} & \textbf{0.674} & \textbf{0.220} \\
						\bottomrule
					\end{tabular}
				}
				\caption{Ablation on the mesh initialization.}
				\label{tab:ablation_init}
			\end{table}
		\end{minipage}
		\begin{minipage}[c]{0.535\textwidth}
			\begin{table}[H]
				\centering
				\resizebox{\linewidth}{!}{
					\begin{tabular}{@{}lccc@{}}
						\toprule
						\ \ Texture Model        & PSNR $\uparrow$ & SSIM $\uparrow$ & LPIPS $\downarrow$ \\ \midrule
						\ \ SI-ViewWarp \cite{tulsiani2020implicit} & 17.81 & 0.540 & 0.318 \\
						\ \ MI-ViewWarp & 19.22 & 0.570 & 0.277 \\
						\ \ Per-vertex Color & 19.37 & 0.595 & 0.270  \\
						\ \ Texture UV Map & 19.51 & 0.618 & 0.263 \\
						\midrule
						\ \ Phong Model \cite{phong1975illumination} & 20.13 & 0.645 & 0.248 \\
						\ \ Cook-Torrance Model \cite{walter2007microfacet} & \textbf{21.72} & \textbf{0.674} & \textbf{0.220} \\
						\bottomrule
					\end{tabular}
				}
				\caption{Ablation on varied texture models.}
				\label{tab:ablation_tex}
			\end{table}
		\end{minipage}
	\end{minipage}
	\vspace{-0.2in}
\end{figure}

\subsection{Robustness of CADSim to Data Noise}
\label{sec:data_noise}
\begin{figure}[t]
	\begin{center}
		\includegraphics[width=0.98\textwidth]{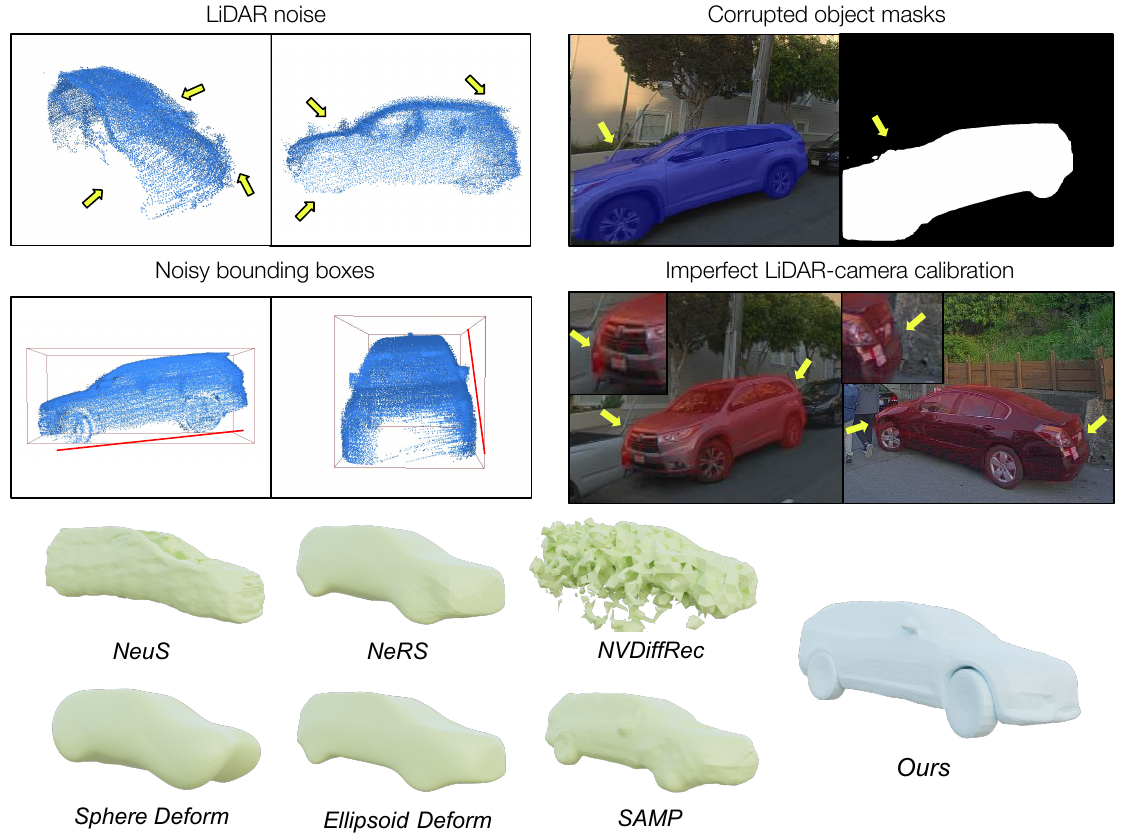}
	\end{center}
	\cuthalfcaptionup
	\caption{
		\textbf{Visualizations of data noise on PandaVehicle.} Sources of noise include (clockwise from top-left): incomplete and noisy LiDAR scans, corrupted object masks, noisy actor bounding boxes and imperfect lidar-camera calibration. 
		Reconstruction with this in-the-wild data is challenging as (a:left) Sparse and incomplete LiDAR points can result in collapsed shapes on the invisible side. (a:right) Noisy LiDAR aggregation will lead to inaccurate or non-smooth geometry surface; (b) The corrupted object masks due to inaccurate segmentation predictions can result in inaccurate geometry and appearance; (c) The lidar-camera calibration errors will result in imperfect geometry that does not align with the images well and potentially has blurry appearance. (d) The noise in bounding boxes will lead to blurry appearance and imperfect initialization and alignment. Compared to existing approaches, CADSim is robust to those data noise meanwhile maintaining fine-grained geometry details and editable parts.
	}
	\cuthalfcaptiondown
	\label{fig:data_noise}
\end{figure}

CADSim is designed to handle real-world driving data captured by SDVs that inevitably contains noise. Sources of noise include: incomplete LiDAR scans, corrupted object masks, noisy actor poses and imperfect lidar-camera calibration. Examples of noise in PandaVehicle data are shown in Figure~\ref{fig:data_noise}. Existing approaches do not perform as well on sparse and noisy in-the-wild data as shown in Figure~\ref{fig:data_noise} and Table 1. Our approach achieves stronger results in this setting qualitatively and quantitatively.

\begin{figure}[t]
	\begin{center}
		\includegraphics[width=0.98\textwidth]{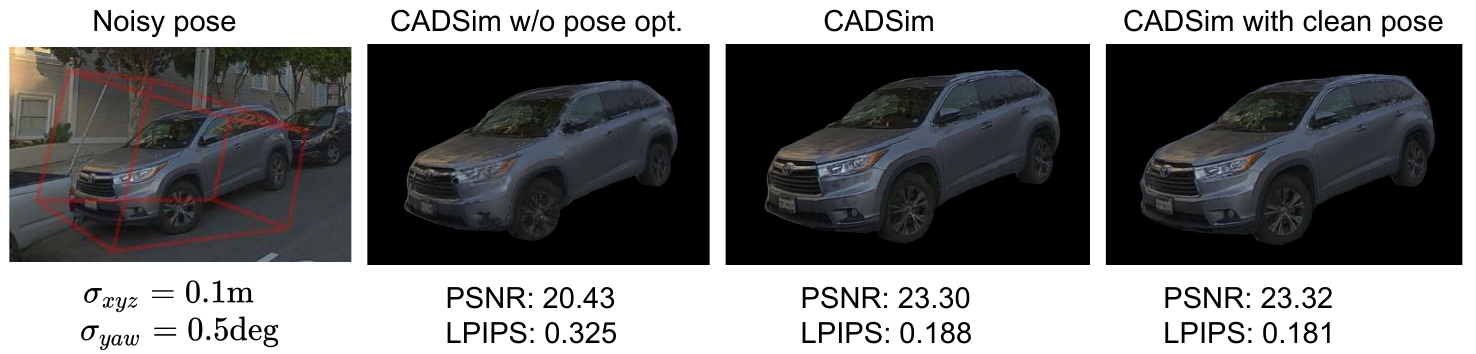}
	\end{center}
	\cuthalfcaptionup
	\caption{\textbf{Ablation study on the pose optimization for CADSim.} From left to right, we show (a) noisy actor bounding box $\xi$ (red) by inserting synthetic Gaussian noise; (b) CADSim learning using perturbed actor poses but with the actor pose optimization module disabled; (c) CADSim learning with noisy poses; (d) CADSim learning using clean actor poses. We can observe the effectiveness of pose optimization in CADSim for obtaining non-blurry rendering results.
	}
	\label{fig:actor_pose_noise}
\end{figure}

To further demonstrate the robustness of CADSim, we conduct the following additional experiments. 

\paragraph{Robustness to actor poses.}
To investigate the robustness of our approach on the noise in the actor poses (3D bounding boxes), we add synthetic noise to the bounding box annotations in PandaVehicle. Specifically, we apply zero-mean Gaussian noise to the center location of the bounding boxes $(x, y, z)$ with variance increasing from 0m to 0.5m. We also insert zero-mean Gaussian noise with variance of 5 degree to the yaw angle. The results for actor \texttt{1d79eded-2fb0-4f89-ba35-323926f45ade} are presented in Table~\ref{tab:actor_pose_noise}. Compared to Instant-NGP~\cite{muller2022instant}, our reconstruction performance only drops slightly as actor pose noise increases. This is because our framework also jointly optimizes the actor pose (\ie, $\xi$ in Eqn. (4)). Note that for Instant-NGP, we also enable the camera extrinsics optimization. Finally, we disable the actor pose optimization component and re-run the experiment. Qualitative results in Figure~\ref{fig:actor_pose_noise} show the effectiveness of pose optimization in obtaining non-blurry rendering results. 

\begin{table}[htbp!]
	\centering
	\resizebox{0.8\textwidth}{!}{
		\begin{tabular}{@{}clccc@{}}
			\toprule
			Noise ($\mu = 0$)             &  Method           & PSNR $\uparrow$ & SSIM $\uparrow$ & LPIPS $\downarrow$                \\ \midrule
			\multirow{2}{*}{$\sigma_{xyz} = 0\mathrm m$} & CADSim      &  23.32 & 0.761 & 0.181 \\
			& Instant-NGP~\cite{muller2022instant} & 22.66 & 0.759 & 0.223   \\ \midrule
			\multirow{2}{*}{$\sigma_{xyz} = 0.1\mathrm m$} & CADSim      &  23.30 \small{\textcolor{ForestGreen}{(-0.02)}} & 0.757 \small{\textcolor{ForestGreen}{(-0.004)}} & 0.188 \small{\textcolor{ForestGreen}{(+0.007)}} \\
			& Instant-NGP~\cite{muller2022instant} & 20.26 \small{\textcolor{purple}{(-2.40)}} & 0.661 \small{\textcolor{purple}{(-0.098)}} & 0.439 \small{\textcolor{purple}{(+0.216)}} \\\midrule
			\multirow{2}{*}{$\sigma_{xyz} = 0.2\mathrm m$} & CADSim      & 23.12 \small{\textcolor{ForestGreen}{(-0.20)}} & 0.744 \small{\textcolor{ForestGreen}{(-0.017)}} & 0.195 \small{\textcolor{ForestGreen}{(+0.014)}}              \\
			& Instant-NGP~\cite{muller2022instant} & 19.77 \small{\textcolor{purple}{(-2.89)}} & 0.655 \small{\textcolor{purple}{(-0.104)}} & 0.461 \small{\textcolor{purple}{(+0.238)}} 
			\\ \midrule
			\multirow{2}{*}{$\sigma_{xyz} = 0.5\mathrm m$} & CADSim      & 22.91 \small{\textcolor{ForestGreen}{(-0.41)}} & 0.725 \small{\textcolor{ForestGreen}{(-0.036)}} & 0.230 \small{\textcolor{ForestGreen}{(+0.049)}} \\
			& Instant-NGP~\cite{muller2022instant} & 19.19 \small{\textcolor{purple}{(-3.47)}} & 0.649 \small{\textcolor{purple}{(-0.110)}} & 0.471 \small{\textcolor{purple}{(+0.248)}} \\ \bottomrule
		\end{tabular}
	}
	\vspace{0.05in}
	\caption{Robustness to noisy actor poses.}
	\label{tab:actor_pose_noise}
	\vspace{-0.15in}
\end{table}

\paragraph{Robustness to corrupted object mask.}
To evaluate the impact of corrupted masks, following~\cite{munkberg2021extracting}, we add synthetic noise into the predicted object masks. Specifically, we find the contour of the object and perturb each contour point pixel location with a zero-mean Gaussian noise. The variance is set from $0 \mathrm{px}$ to $50 \mathrm{px}$. We train the model using the front camera and corrupted object masks and test on the front-left camera for the actor \texttt{1d79eded-2fb0-4f89-ba35-323926f45ade}.
Experiments in Table~\ref{tab:corrupted_obj_mask} show that our approach is quite robust to the corrupted masks even in a high-noise regime (e.g., $\sigma = 50 \mathrm{px}$) thanks to the robust CAD initialization and our carefully designed energy formulation.

\begin{table}[htbp!]
	\centering
	\resizebox{0.65\textwidth}{!}{
		\begin{tabular}{@{}cccc@{}}
			\toprule
			Noise ($\mu = 0$)                        & PSNR $\uparrow$ & SSIM $\uparrow$ & LPIPS $\downarrow$                \\ \midrule
			\multirow{1}{*}{$\sigma = 0  \mathrm{px}$}  &  23.32 & 0.761 & 0.181 \\
			\multirow{1}{*}{$\sigma = 5  \mathrm{px}$}  &  23.29 \small{\textcolor{ForestGreen}{(-0.03)}} & 0.756 \small{\textcolor{ForestGreen}{(-0.005)}} & 0.192 \small{\textcolor{ForestGreen}{(+0.011)}} \\
			\multirow{1}{*}{$\sigma = 10  \mathrm{px}$} & 23.22 \small{\textcolor{ForestGreen}{(-0.10)}} & 0.753 \small{\textcolor{ForestGreen}{(-0.008)}} & 0.198 \small{\textcolor{ForestGreen}{(+0.017)}}   \\
			\multirow{1}{*}{$\sigma = 20  \mathrm{px}$} & 22.99 \small{\textcolor{ForestGreen}{(-0.34)}} & 0.743 \small{\textcolor{ForestGreen}{(-0.018)}} & 0.214 \small{\textcolor{ForestGreen}{(+0.023)}} \\
			\multirow{1}{*}{$\sigma = 50  \mathrm{px}$} & 22.56 \small{\textcolor{ForestGreen}{(-0.66)}} & 0.737 \small{\textcolor{ForestGreen}{(-0.024)}} & 0.223 \small{\textcolor{ForestGreen}{(+0.032)}} \\
			\bottomrule
		\end{tabular}
	}
	\vspace{0.05in}
	\caption{Robustness to corrupted object masks.}
	\label{tab:corrupted_obj_mask}
	\vspace{-0.15in}
\end{table}

In summary, we believe that our approach is more robust to data noise and recovers more accurate geometries from sparse/noisy data compared to existing approaches.

\subsection{Reconstruction of Non-Vehicle Classes}
\label{sec:other_classes}

While we mainly focus on the vehicle reconstruction as vehicles are the primary
traffic participants, we note that CAD models are readily available for most object classes, and that our approach can be extended to other
classes. In this section, we showcase the reconstruction of motorcycles and cones on PandaSet using CADSim.

\begin{figure}[htbp!]
	\begin{center}
		\includegraphics[width=0.9\textwidth]{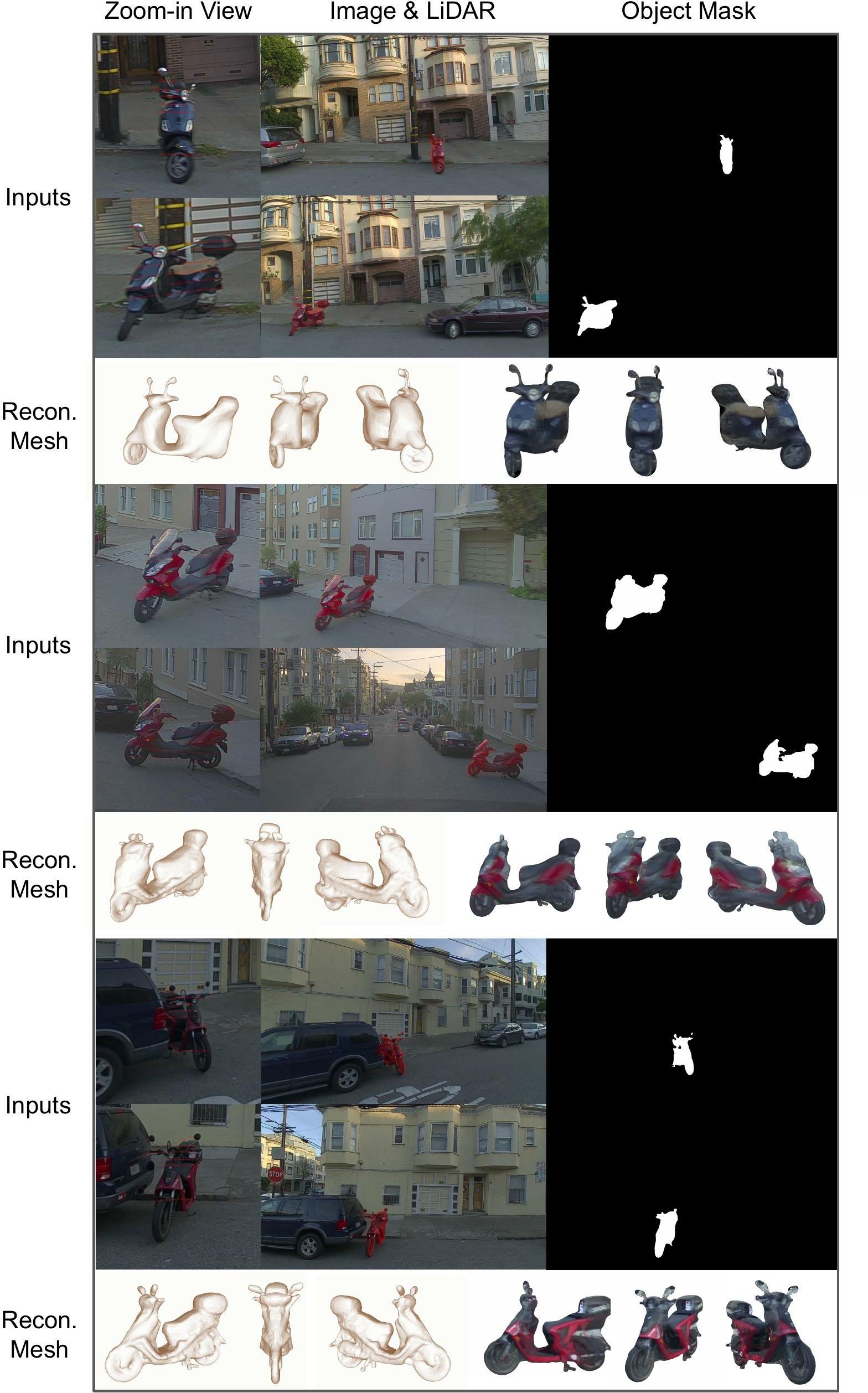}
	\end{center}
	\cuthalfcaptionup
	\caption{Reconstruction of motorcycles on PandaSet.}
	\label{fig:motorcycle}
\end{figure}

\paragraph{Challenges:} Motorcycles are more challenging to reconstruct in-the-wild. Since they are smaller than vehicles, the scanned lidar points are sparser and noisier, and there are fewer image pixels corresponding to the motorcycles. Moreover, due to complex topology and occlusion, the object masks predicted by PointRend are more corrupted (see Figure~\ref{fig:motorcycle}) and there are larger relative camera/lidar misalignments for small objects. Please refer to Sec~\ref{sec:additional_limitation} for potential future directions.

\begin{figure}[htbp!]
	\begin{center}
		\includegraphics[width=\textwidth]{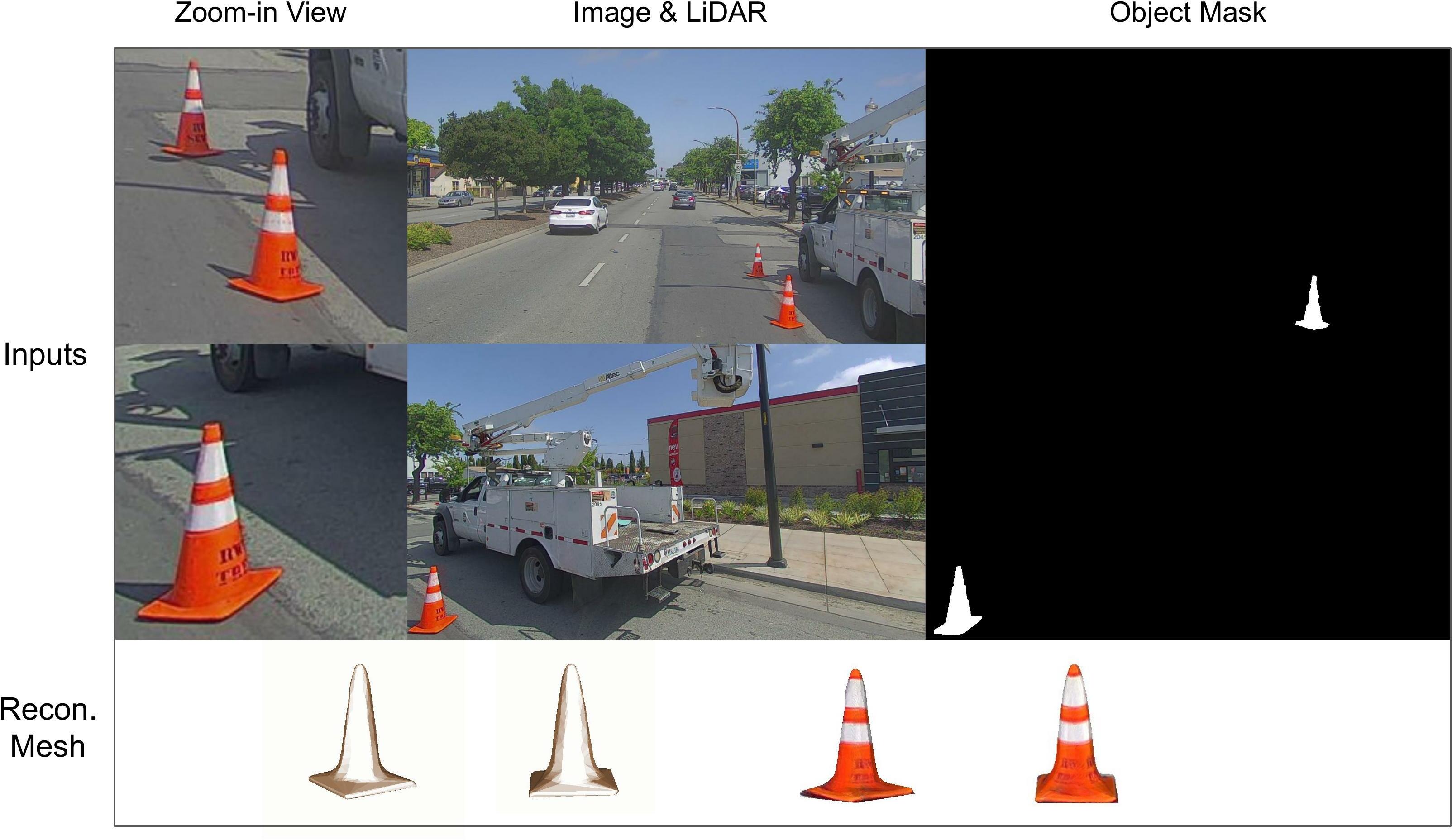}
	\end{center}
	\cuthalfcaptionup
	\caption{Reconstruction of traffic cones on PandaSet.}
	\label{fig:cone}
\end{figure}

We encode the semantic priors for a single CAD asset and obtain three parts (handlebar, front wheel, and the rest body).
We optimize the per vertex offset for the body and handlebar parts. To make motorcycle behave in a physically plausible manner, for front wheel and handlebar, we also optimize the relative pose to the motorcycle origin, scale factors, relative rotation, and translation offset.
As shown in Figure~\ref{fig:motorcycle}, CADSim is able to reconstruct motorcycles with reasonable geometry and appearance in spite of large data noise (\ie, large LiDAR image misalignment shown in the leftmost column, coarse and noisy segmentation mask) and very sparse observations. In Figure~\ref{fig:cone}, we use CADSim to reconstruct a traffic cone, demonstrating that CAD priors enable efficient and accurate mesh reconstruction for very small objects as well.

\subsection{Additional Limitations}
\label{sec:additional_limitation}
In addition to the limitations described in the main paper, we describe additional limitations and possible extensions.
Our method does not allow for arbitrary changes in mesh topology and it might lead to some unexpected behavior when the topology changes within one class (see the windshield in Figure~\ref{fig:motorcycle}), which is an exciting future direction to address. Futhermore, we use aggregated LiDAR points instead of doing per-frame registration.
This may make the approach less robust to LiDAR image misalignment, especially for small or far-away objects.
While \model_name is more robust than other reconstruction approaches, it might still produce unexpected results on extremely noisy data (\eg, large calibration misalignment, highly corrupted object masks).

With regards to potential extensions, rigging additional parts
(\eg, vehicle doors) or even automatic rigging may be conducted in future works. Finally, our experiments also focus on sensor simulation for perception models, and we may investigate the performance for downstream planning tasks for future works.

\subsection{Additional Downstream Evaluation on Camera Simulation}
\label{sec:downstream_eval}

To verify if \model_name helps reduce domain gap for downstream perception tasks consistently, we evaluate another perception model on simulated camera images at novel views. Specifically, we follow the same setting as Table 4 in the main paper except replacing Mask R-CNN~\cite{he2017mask} with PointRend~\cite{kirillov2020pointrend} algorithms.
As shown in Table~\ref{tab:downstream_evaluation_supp}, using \model_name assets usually leads to the largest agreement with real images under different settings. This indicates the effectiveness of reconstructed CADSim assets for end-to-end autonomy testing.

\setlength{\tabcolsep}{6.5pt}
\begin{table}[htbp!]
	\centering
	\resizebox{\textwidth}{!}{
		\begin{tabular}{lccccccccccccc}
			\toprule
			& \multicolumn{5}{c}{\textit{Left camera $\rightarrow$ Front-left camera}} & &  \multicolumn{5}{c}{\textit{Left camera $\rightarrow$ Front camera}} \\ \cline{2-6}\cline{8-12}
			& \multicolumn{2}{c}{Blending~\cite{chen2021geosim}} & & \multicolumn{2}{c}{Copy-Paste} & & \multicolumn{2}{c}{Blending~\cite{chen2021geosim}} & & \multicolumn{2}{c}{Copy-Paste} \\
			& \multicolumn{1}{c}{Det.} & \multicolumn{1}{c}{Segm.} & & \multicolumn{1}{c}{Det.} & \multicolumn{1}{c}{Segm.} & & \multicolumn{1}{c}{Det.} & \multicolumn{1}{c}{Segm.} & & \multicolumn{1}{c}{Det.} & \multicolumn{1}{c}{Segm.} \\ \midrule
			Instant-NGP &  86.18 & 87.03 & & 79.84 & 80.61 & & 71.74 & 71.22 & & 50.08 & 49.05 \\
			NeuS & \underline{94.59} & \textbf{95.24} & &  92.83 & \underline{93.72} & & 75.03 & 74.65 &  & 69.47 & 68.91 \\
			SAMP & 90.82 & 90.01 & & 90.88 & 90.85 & & \underline{81.79} & 78.00 & & \textbf{82.39} & \textbf{82.01} \\
			\rowcolor{grey}\model_name & \textbf{94.71} & \underline{94.43} & & \textbf{94.68} & \textbf{94.66}  & &  \textbf{82.68} & \textbf{80.81} & & \underline{82.33} & \underline{81.32} \\
			\bottomrule
		\end{tabular}
	}
	\vspace{0.05in}
	\caption{Evaluation of downstream perception tasks (\ie, object detection, instance segmentation) on camera simulation. We report the metric agreement (instance-level IoU) with the model evaluated on real data.
	}
	\label{tab:downstream_evaluation_supp}
\end{table}

\section{Realistic and Controllable Simulation}
\label{sec:simulation}

\subsection{Log-Replay Simulation with CADSim Assets}
\label{sec:sim_log_replay}
\begin{figure}[htbp!]
	\begin{center}
		\includegraphics[width=\textwidth]{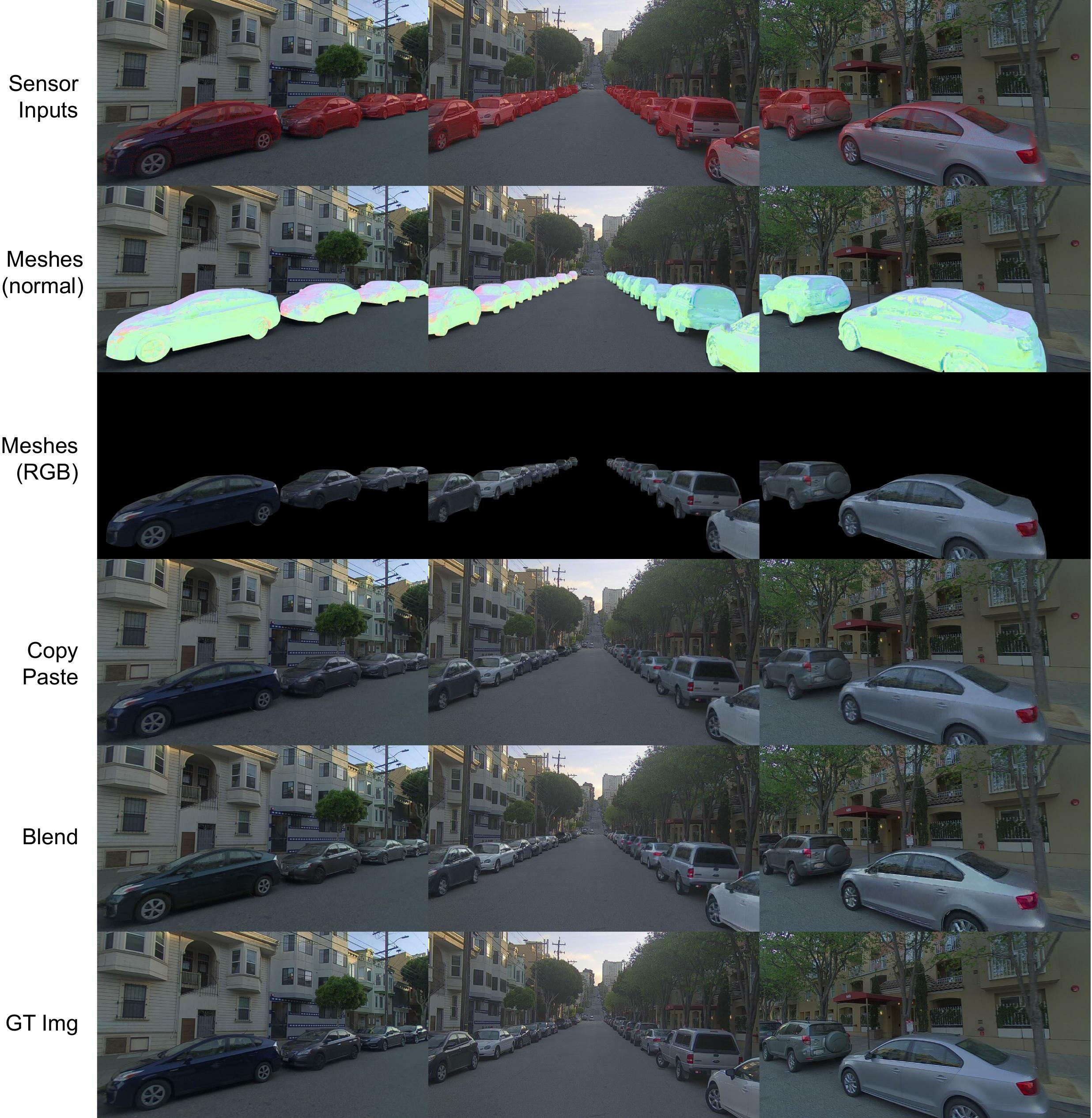}
	\end{center}
	\cuthalfcaptionup
	\caption{Reconstruction of all nearby vehicles and camera re-simulation on PandaSet \texttt{Log028}.}
	\label{fig:log028_camerasim}
\end{figure}

\begin{figure}[htbp!]
	\begin{center}
		\includegraphics[width=0.95\textwidth]{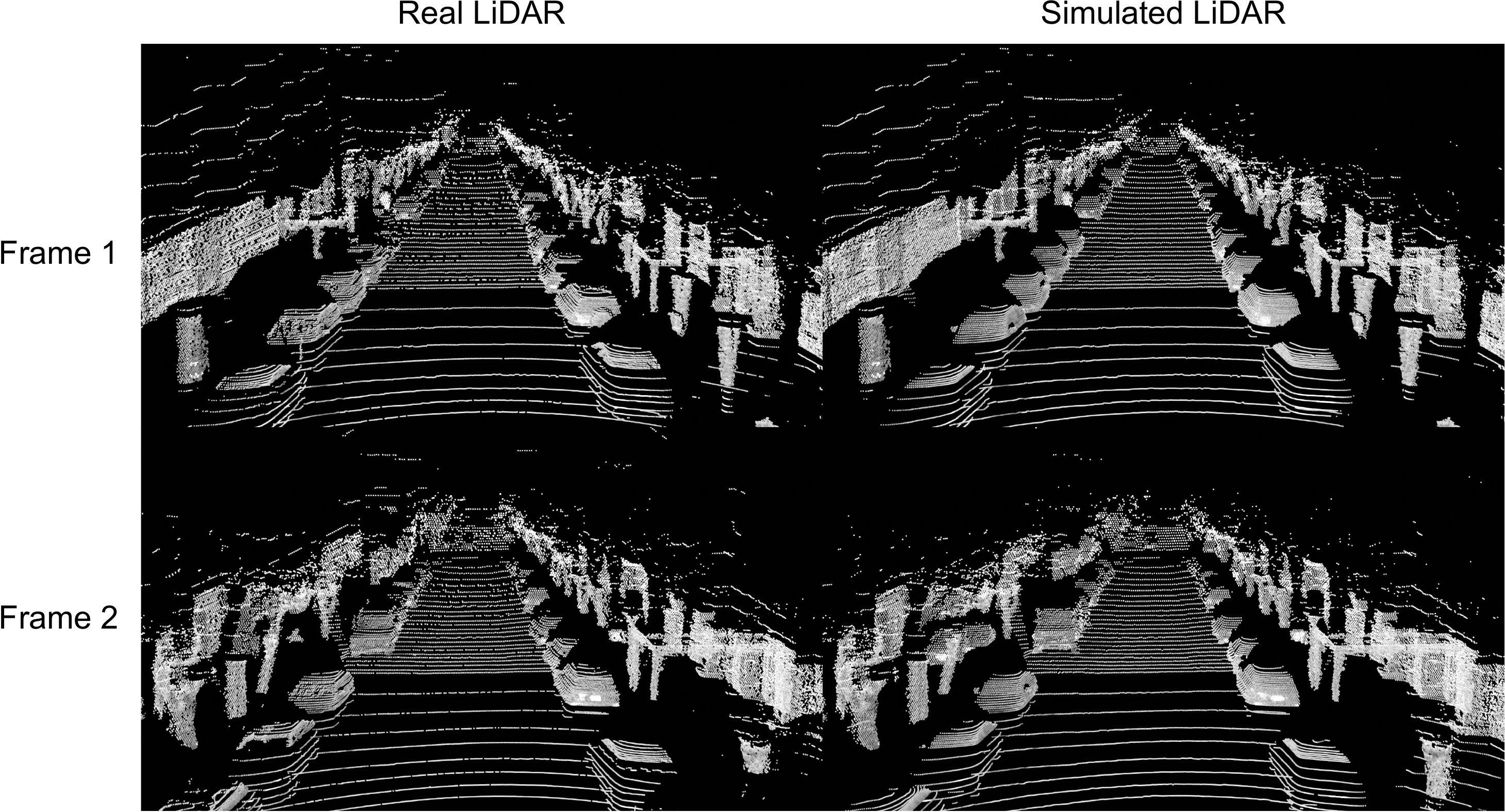}
	\end{center}
	\cuthalfcaptionup
	\caption{LiDAR re-simulation using CADSim assets on PandaSet \texttt{Log028}.}
	\label{fig:log028_lidarsim}
\end{figure}

\begin{figure}[htbp!]
	\begin{center}
		\includegraphics[width=\textwidth]{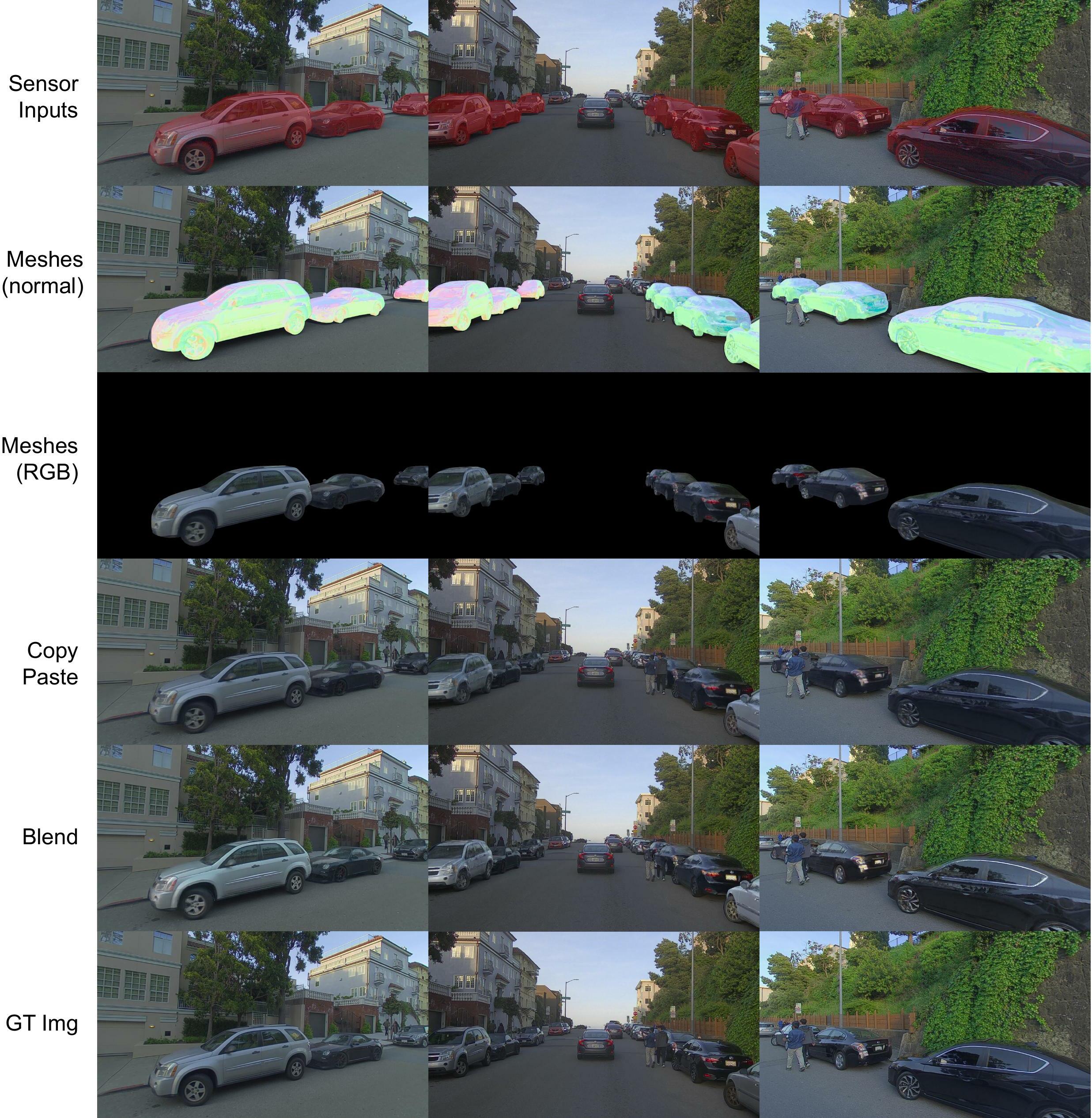}
	\end{center}
	\cuthalfcaptionup
	\caption{Reconstruction of all nearby vehicles and camera re-simulation on PandaSet \texttt{Log030}.}
	\label{fig:log030_camerasim}
\end{figure}

\begin{figure}[htbp!]
	\begin{center}
		\includegraphics[width=0.95\textwidth]{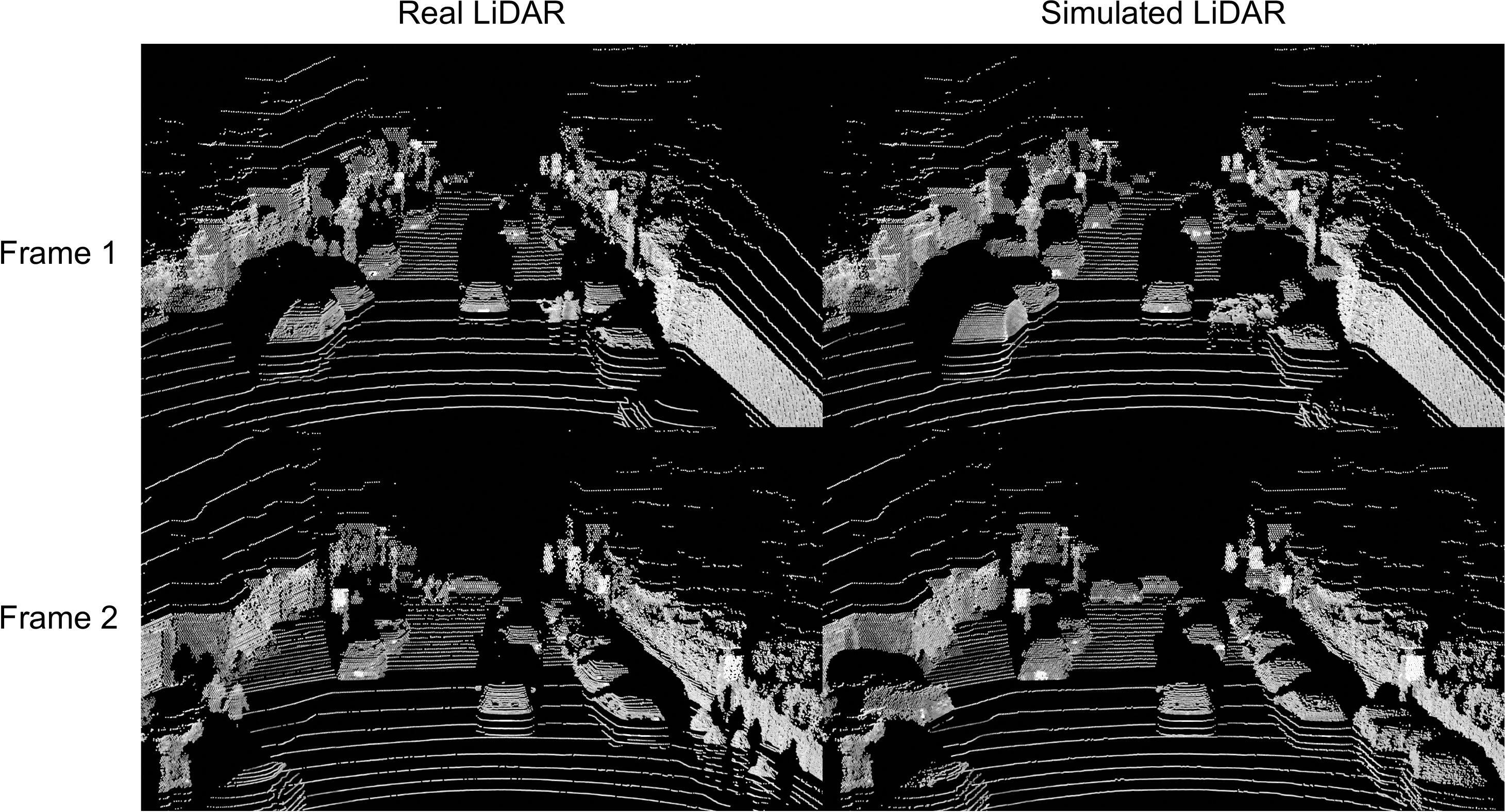}
	\end{center}
	\cuthalfcaptionup
	\caption{LiDAR re-simulation using CADSim assets on PandaSet \texttt{Log030}.}
	\label{fig:log030_lidarsim}
\end{figure}

\begin{figure}[htbp!]
	\begin{center}
		\includegraphics[width=\textwidth]{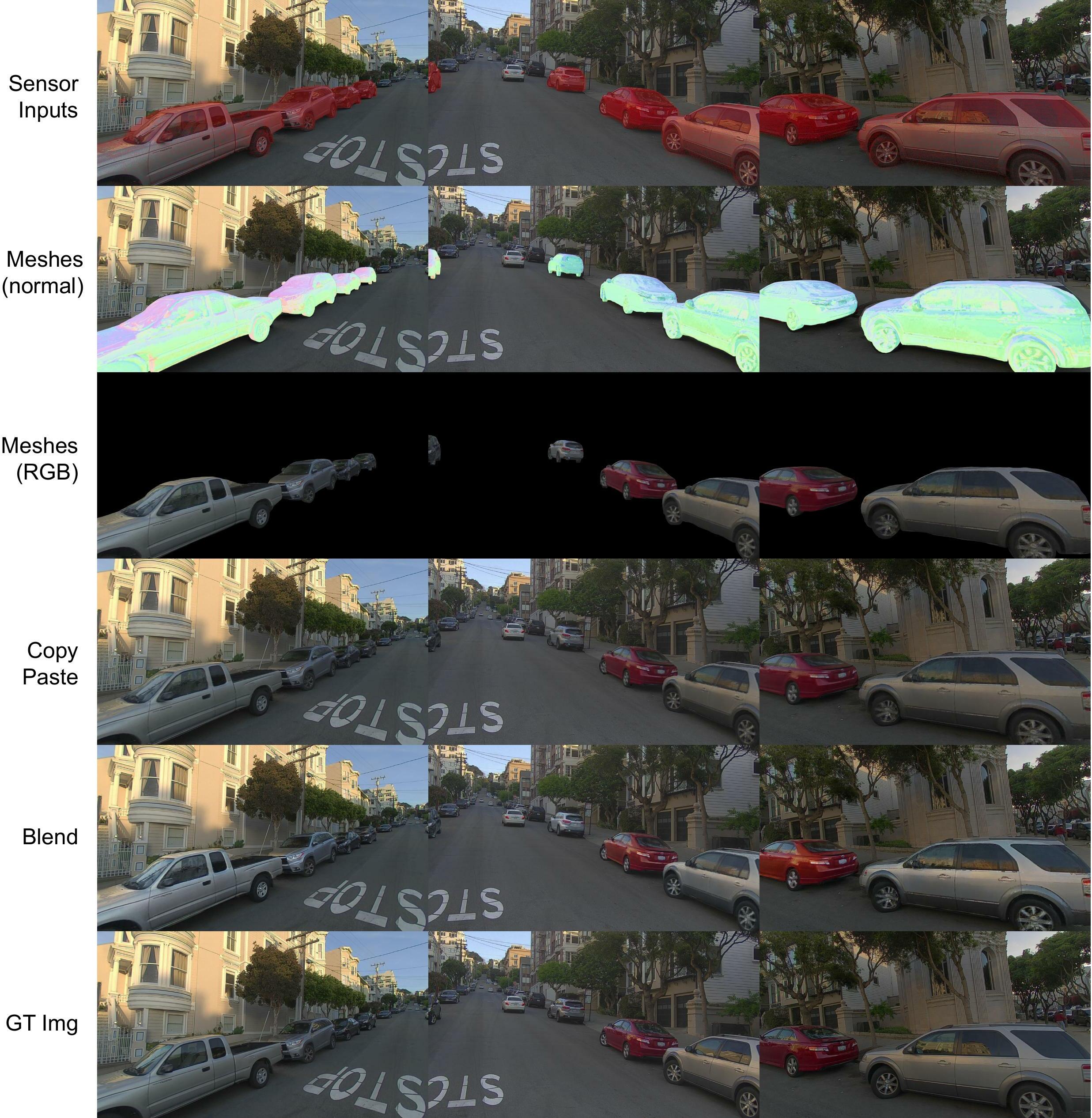}
	\end{center}
	\cuthalfcaptionup
	\caption{Reconstruction of all nearby vehicles and camera re-simulation on PandaSet \texttt{Log139}.}
	\label{fig:log139_camerasim}
\end{figure}

\begin{figure}[htbp!]
	\begin{center}
		\includegraphics[width=0.95\textwidth]{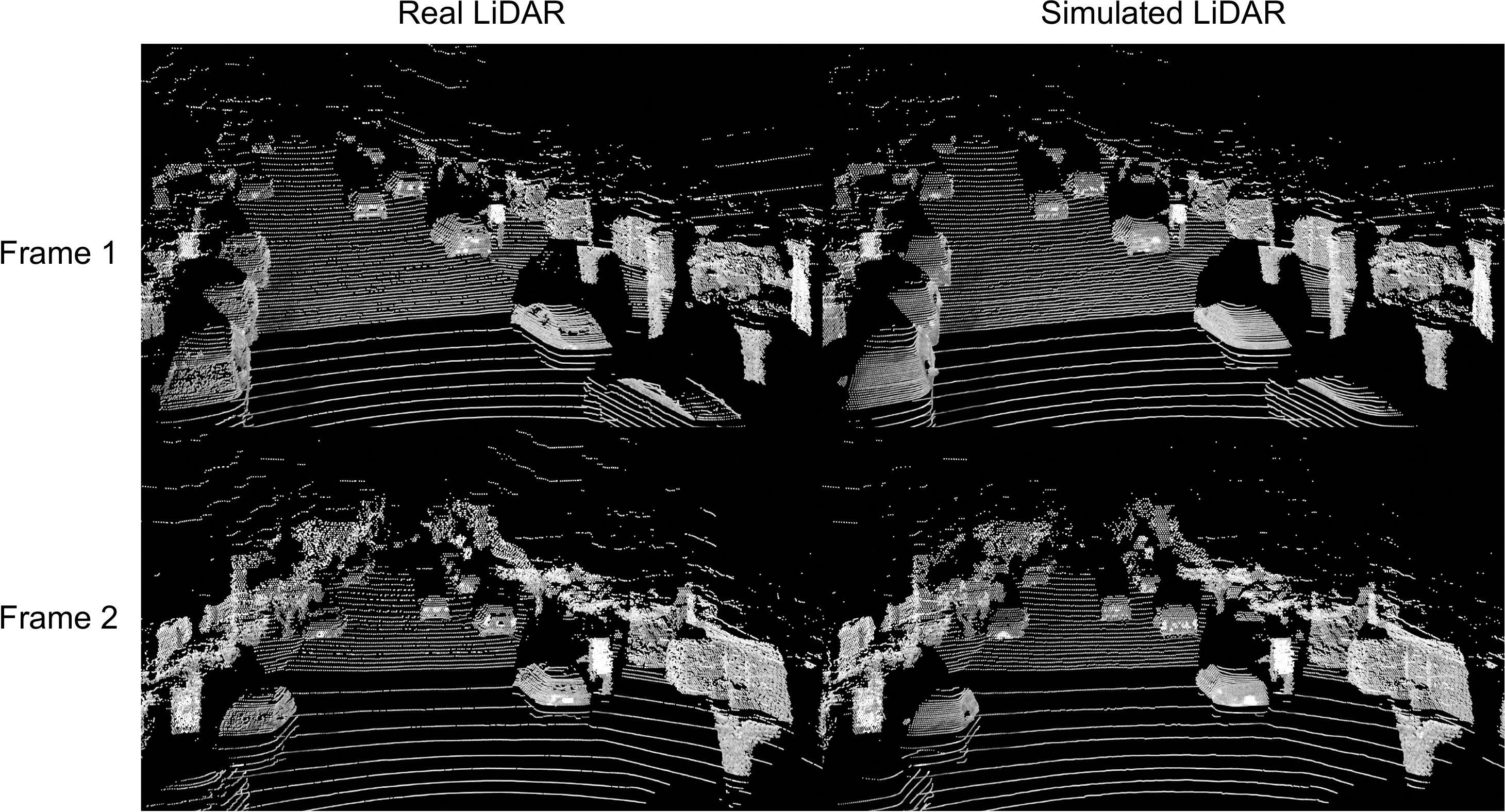}
	\end{center}
	\cuthalfcaptionup
	\caption{LiDAR re-simulation using CADSim assets on PandaSet \texttt{Log139}.}
	\label{fig:log139_lidarsim}
\end{figure}

We now show that CADsim can reconstruct the static vehicles at scale from sequences of sensor data.
In Figures~\ref{fig:log028_camerasim}-\ref{fig:log139_lidarsim} we show the original sensor data for both camera and LiDAR, the reconstructed meshes with CADsim rendered by normal and RGB appearance, and then show the simulated sensor data generated when placing the assets in their original views.
For camera simulation, we follow GeoSim~\cite{chen2021geosim},
where given a scenario configuration, we first render the asset to the target view and then apply a post composition network and conduct occlusion reasoning to seamlessly blend the actor to real backgrounds.
For LiDAR simulation, similar to~\cite{manivasagam2020lidarsim,wang2021advsim}, we use a high-fidelity Pandar64~\cite{xiao2021pandaset} simulator that conducts LiDAR ray-casting on the added actor according to the LiDAR calibration and remove
points in the real LiDAR sweep that are occluded by the added actor.
CADsim reliably constructs the nearby vehicles with high quality geometry and appearance, and have high fidelity with the original sensor data. These assets thereby provide a more accurate way of generating sensor data for simulation scenarios.

\subsection{Mixed Reality with Realistic Animated Vehicle Insertion}
\label{sec:mixed_reality}
The previous section demonstrated \model_name generates assets that match with the original sensor data. We can now insert these assets into existing scenarios and create new variations.
As we have accurate 3D meshes, we can also render shadows for the inserted actor.
	We use the rasterization engine PyRender \cite{matl2019pyrender} to generate the shadow based on the geometry of the inserted actor, assuming a top-down light.

\begin{figure}[htbp!]
	\hspace{-0.55in}
	\includegraphics[height=0.8\textheight]{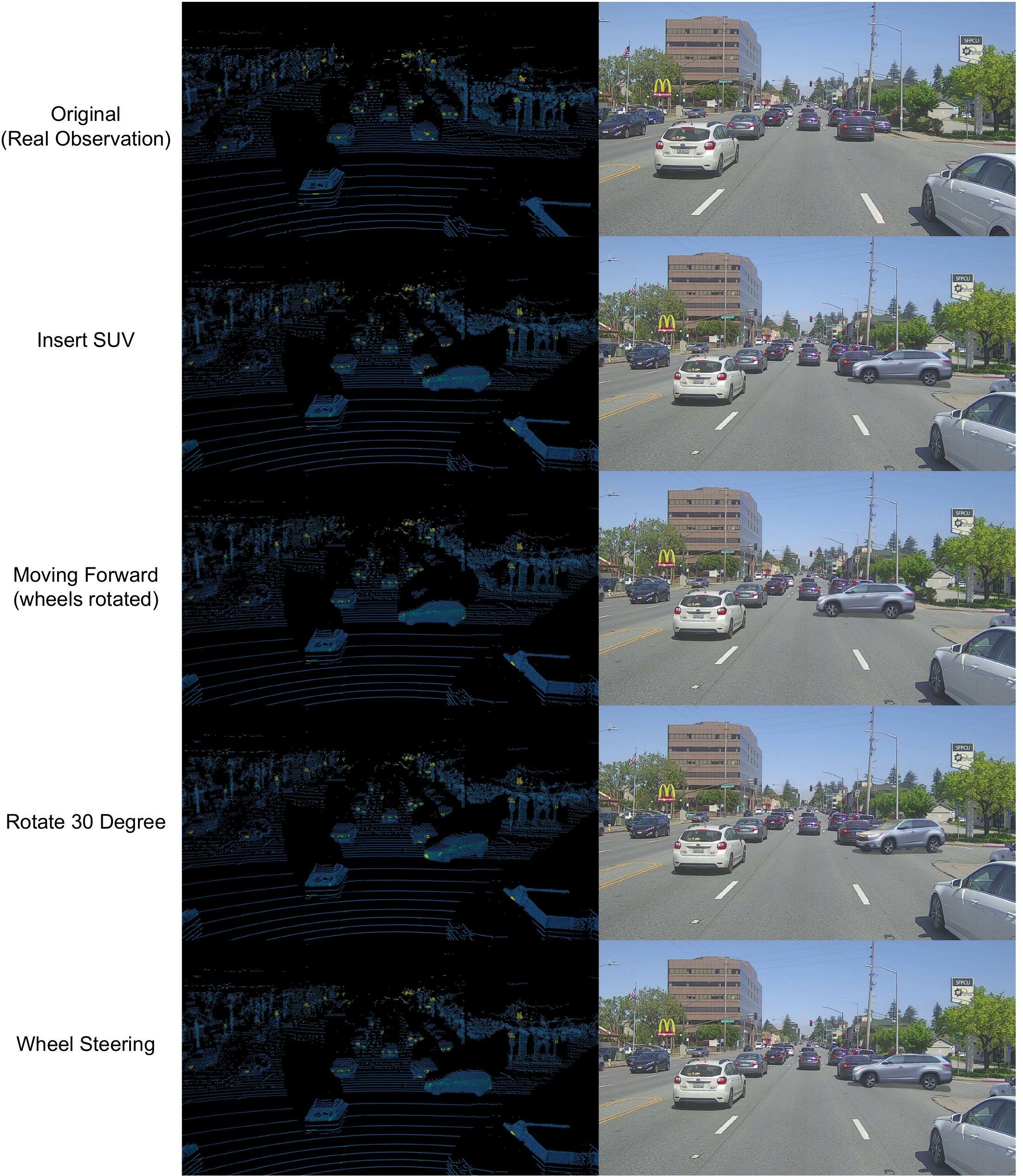}
	\cuthalfcaptionup
	\caption{Mixed reality actor manipulation with articulated CADSim vehicle.}
	\label{fig:scene_manipulate}
\end{figure}

We now show how CADsim assets can be inserted into new scenes and can be manipulated to create variations. As shown in Figure~\ref{fig:scene_manipulate}, we render the CADsim asset at a new pose and blend the rendered image segment into the new scene. The position and rotation of the added car are manipulated. Since our assets are multi-sensor consistent, we can generate the LiDAR point cloud (left)  and the camera image (right) for the modified scene. Both simulated sensor data look realistic. We notice that the wheels spin when moving forward and backward, and the wheels rotating when model turning.

\begin{figure}[htbp!]
	\hspace{-0.3in}
	\includegraphics[width=1.1\textwidth]{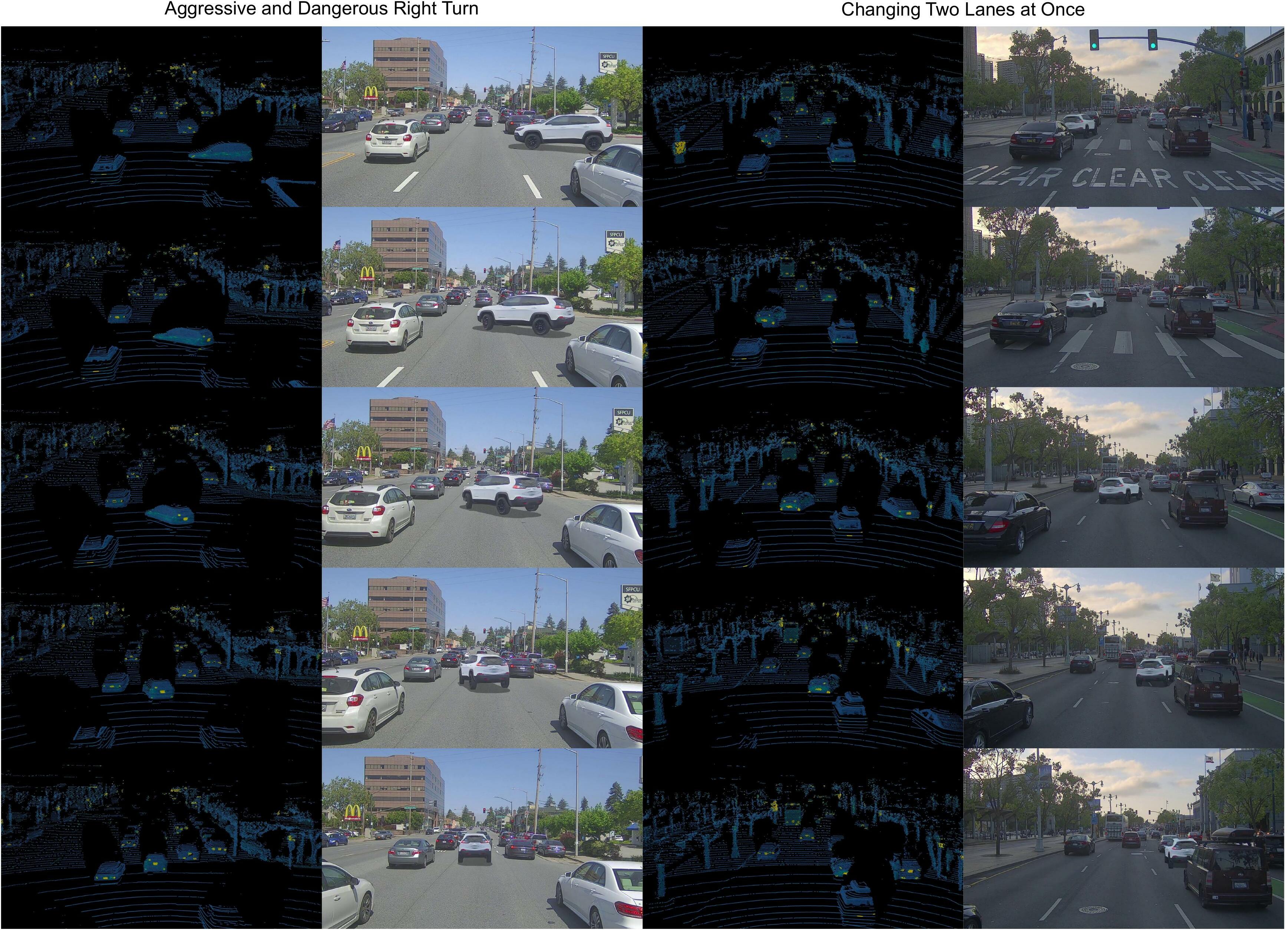}
	\cuthalfcaptionup
	\caption{Realistic multi-sensor simulation for safety-critical mixed reality scenarios.}
	\label{fig:safety_critical}
\end{figure}

\begin{figure}[htbp!]
	\hspace{-0.3in}
	\includegraphics[width=1.1\textwidth]{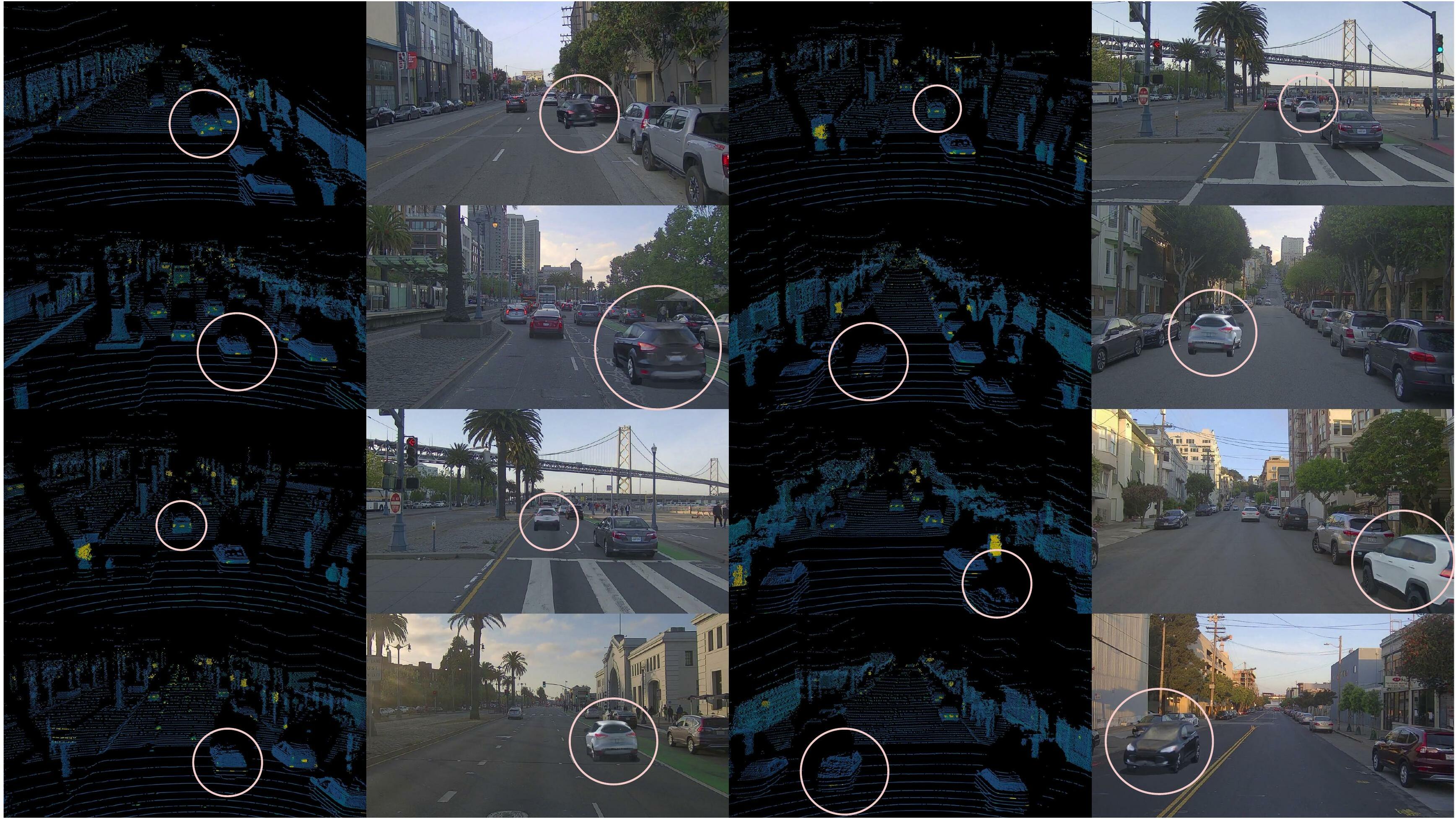}
	\cuthalfcaptionup
	\caption{Realistic multi-sensor actor insertion simulation.}
	\label{fig:actor_insertion_grid}
\end{figure}

We can also generate interesting new scenarios with CADSim assets to test our autonomy systems. In Figure~\ref{fig:safety_critical}, we generate two safety-critical scenarios (left: an actor aggressively turning right into our lane; right: a moving vehicle aggressively changing two lanes at once) and show realistic image and LiDAR simulation. The simulated camera and LiDAR data are blended seamlessly into the original scenario, creating more interesting long-tail scenarios. Note that the occlusion and actor movement are physically plausible. We provide more examples of actor insertion in Figure~\ref{fig:actor_insertion_grid}.

\begin{figure}[htbp!]
	\hspace{-0.4in}
	\includegraphics[width=1.1\textwidth]{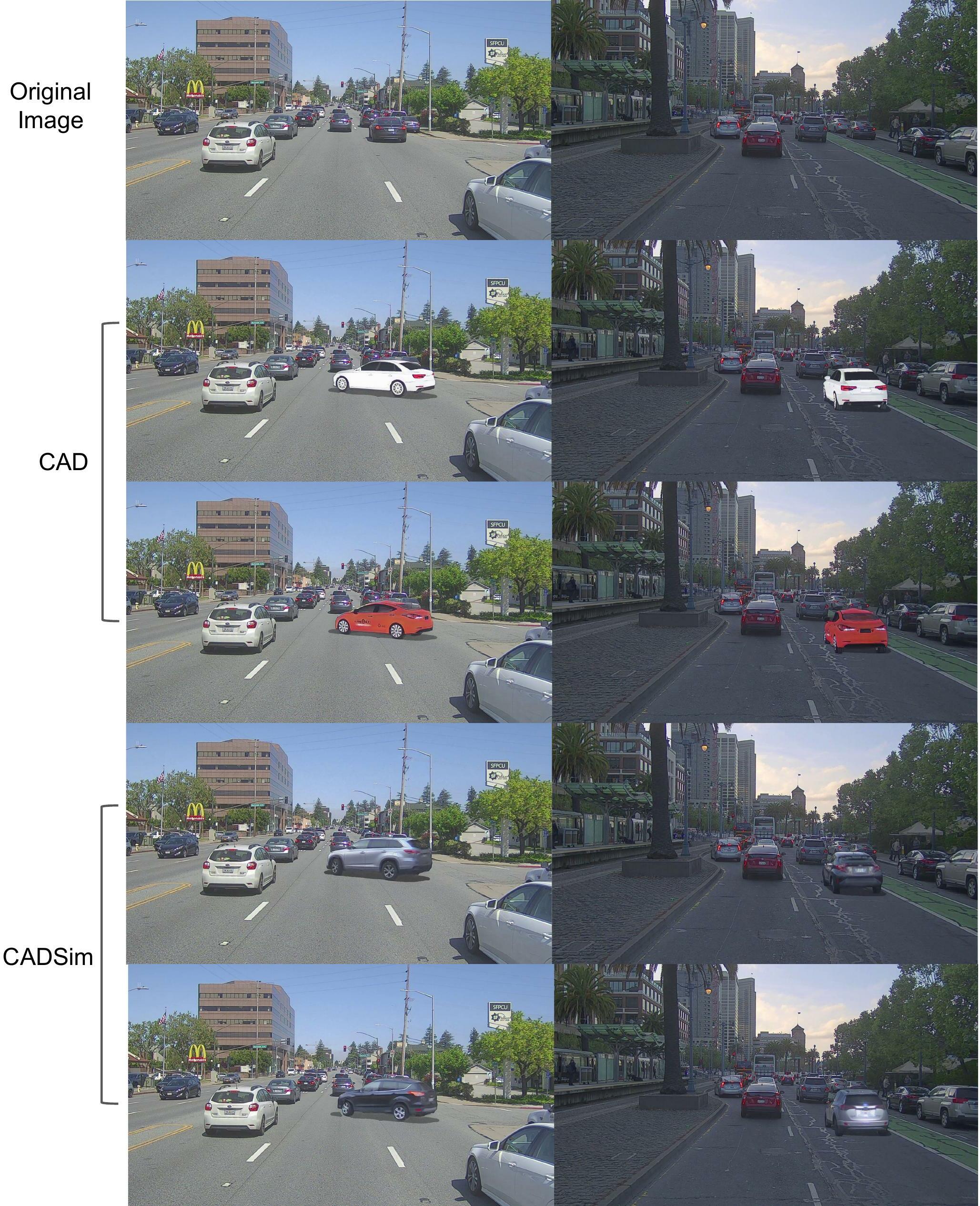}
	\cuthalfcaptionup
	\caption{Comparison of camera actor insertion with CAD or CADSim assets. CADSim assets have more realistic textures and blend in better with the real image. CADSim enables generation of sensor data for completely new scenarios.}
	\label{fig:cad_compare}
\end{figure}

Finally, we compared to directly inserting a CAD model into the original scenario (Figure~\ref{fig:cad_compare}), our inserted CADsim assets have more realistic appearance and harder to distinguish as simulated.

\subsection{Texture Transfer and Synthesis}
\label{sec:tex_transfer}

\begin{figure}[htbp!]
	\hspace{-0.3in}
	\includegraphics[width=1.1\textwidth]{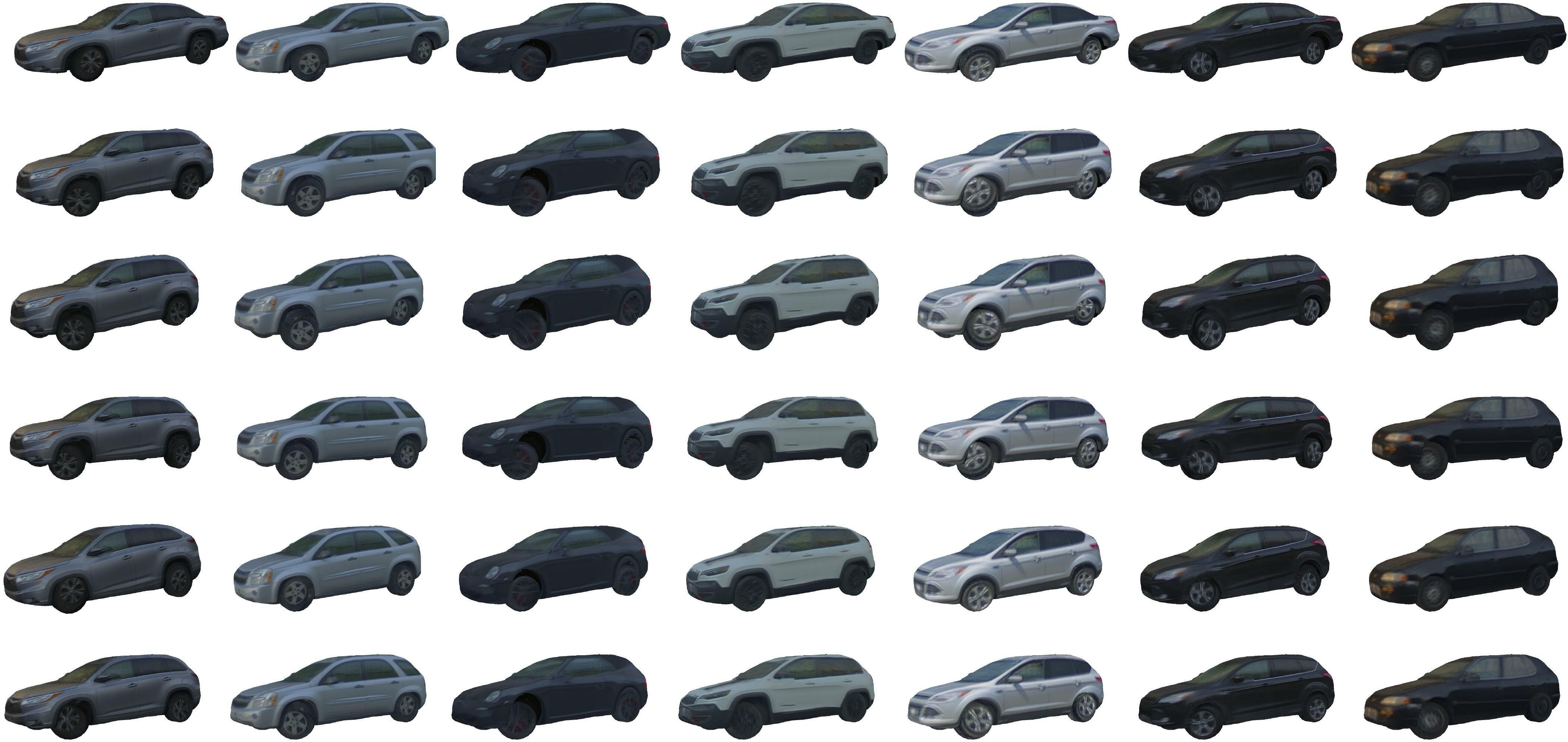}
	\cuthalfcaptionup
	\caption{Texture transfer for reconstructed real-world assets.}
	\label{fig:texture_transfer_assets}
\end{figure}

\begin{figure}[htbp!]
	\hspace{-0.42in}
	\includegraphics[width=1.1\textwidth]{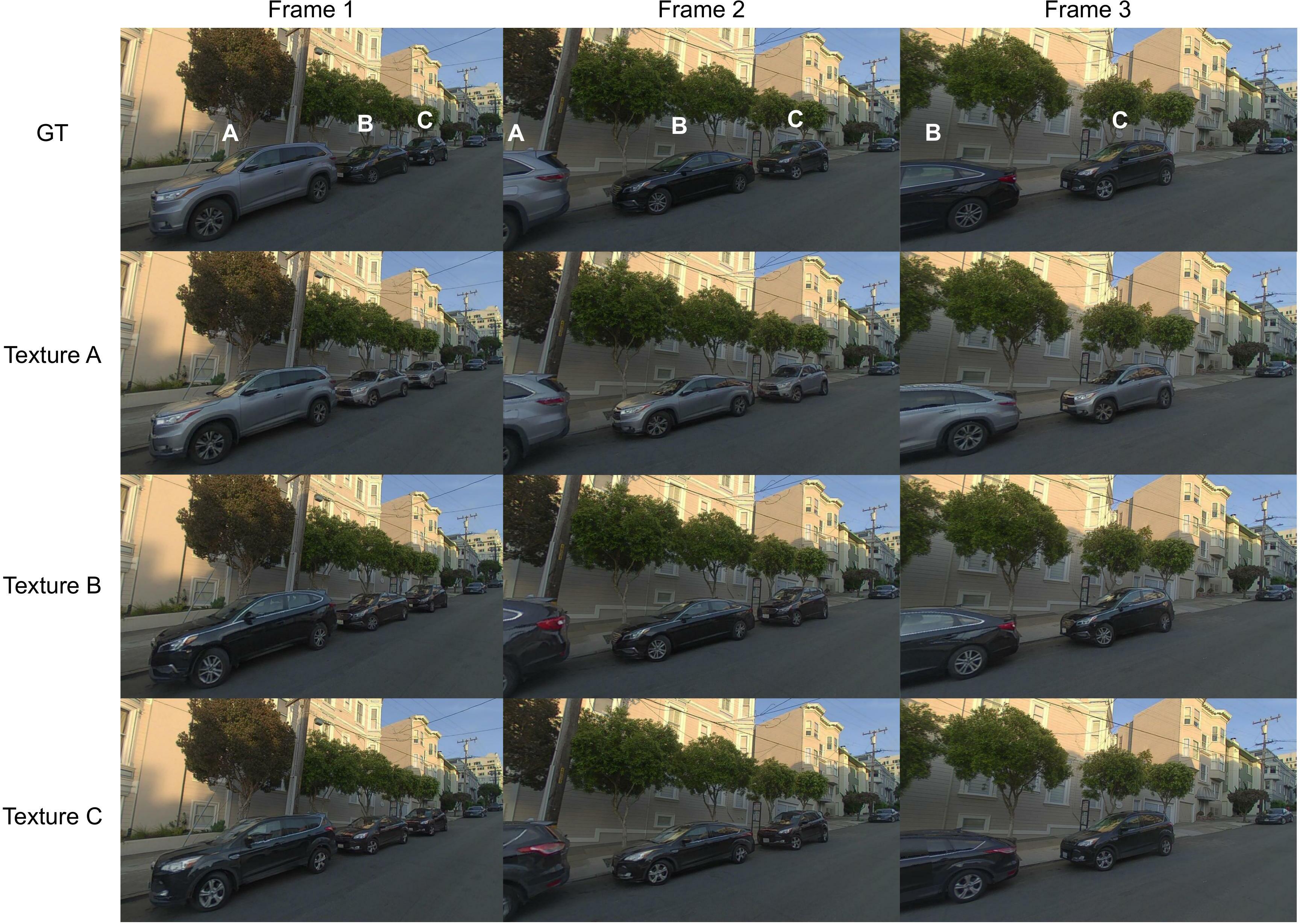}
	\cuthalfcaptionup
	\caption{Swapping vehicle textures in the real world.}
	\label{fig:texture_transfer}
\end{figure}

Finally, we show that our approach can align textures across
different vehicle shapes, enabling texture transfer to create new asset variations.
Unlike recent work (\eg, AUV-Net~\cite{chen2022AUVNET}) that focuses on synthetic objects with unrealistic textures, we demonstrate the texture transfer across multiple actors in the real world and use it for realistic camera simulation.
Figure~\ref{fig:texture_transfer_assets} shows in each column the texture trasferred across different vehicle types, while each row shows the same vehicle shape having different textures.
In Figure~\ref{fig:texture_transfer}, we choose three nearby actors and transfer the texture from one actor to the other ones. Our assets are vertex-aligned with high-quality part correspondence across different shapes, allowing us for realistic and seamless simulation. This technique can also be used to generate diverse textured assets that have never been observed.

\begin{table}[htbp!]
	\centering
	\resizebox{\textwidth}{!}{
		\begin{tabular}{cc c c c c}
			\toprule
			& & \multicolumn{1}{c }{\bf Training view frame id} & \phantom{x} & \multicolumn{2}{c}{\bf Testing view frame id}\\
			\midrule
			\textbf{Actor UUID} & \textbf{Seq id} & \textbf{Left camera} & & \textbf{Front camera} &
			\textbf{Front-left camera}\\
			\midrule
			1d79eded-2fb0-4f89-ba35-323926f45ade & 139 & 46-63 & & - & 44-55\\
			\midrule
			f7bd1486-1fbe-4f33-ba28-f00dae3e0298 & 139 & 57-77 & & 38-53 & 54-69\\
			\midrule
			526e2f5e-e294-415c-aad6-578d27921465 & 030 & 38-78 & & 0-28 & 35-55\\
			\midrule
			56e10a51-35ed-43b0-837c-cea8aff216cc & 139 & 26-52 & & - & 25-46\\
			\midrule
			ba222d39-2f13-4849-8ff4-91e247d5cedf & 120 & 12-37 & & 0-6 & 0-25\\
			\midrule
			2160d735-3fda-49f8-9bd9-e2cba3b51faa & 038 & 34-47 & & 0-30 & 27-41\\
			\midrule
			1be68ce6-68c5-467f-abb1-fa5e03d1db7a & 053 & 33-36, 40-49 & & 0-29 & 25-41\\
			\midrule
			2ee4d8f8-af0a-48f3-bb6c-ed479a7829e7 & 039 & 47-67 & & 0-44 & 28, 31-59\\
			\midrule
			94c06b25-d17a-4ee7-a2df-7faa619bee89 & 035 & 49-58, 60-61 & & 4, 7, 9-42 & 47-51\\
			\midrule
			5ce5fb69-038d-4f82-8c64-90b73c6f6681 & 030 & 17-62 & & 0-17 & 0-45\\
			\bottomrule
		\end{tabular}
	}
	\vspace{0.05in}
	\caption{Actors and camera image frame ids used in PandaVehicle. `-' means no images are available for the associated camera.}
	\label{tab:pandaset-actors}
\end{table}

\end{document}